\documentclass{article}
\usepackage{subfigure,algorithm}
\usepackage{amsmath,amssymb}
\usepackage{graphics,epsfig}
\usepackage{subfigmat}

\usepackage{overpic}
\usepackage{arxiv}
\usepackage[utf8]{inputenc} 
\usepackage[T1]{fontenc}    
\usepackage{url}            
\usepackage{booktabs}       
\usepackage{amsfonts}       
\usepackage{nicefrac}       
\usepackage{microtype}      
\usepackage{lipsum}		
\usepackage{graphicx}
\usepackage{natbib}
\usepackage{xcolor}

\newcommand{\RR}[1]{\ensuremath{\mathbb{R}^{ #1 }}}
\newcommand{\RRstar}[1]{\ensuremath{\mathbb{R}_\star^{ #1 }}}

\newcommand{\solution}{u}
\newcommand{\error}{\boldsymbol e}
\newcommand{\erroro}{\boldsymbol e_o}
\newcommand{\errori}{\boldsymbol e_i}
\newcommand{\errorFOM}{\boldsymbol e_\textit{FOM}(t)}
\newcommand{\errorROM}{\boldsymbol e_\textit{ROM}(t)}
\newcommand{\errorFOMave}{\boldsymbol e_\textit{FOM}}
\newcommand{\errorROMave}{\boldsymbol e_\textit{ROM}}
\newcommand{\tauFOM}{\tau_\textit{FOM}}
\newcommand{\tauROM}{\tau_\textit{ROM}}
\newcommand{\mystateGalerkin}{\hat {\boldsymbol x_{b}}}
\newcommand{\mystate}{\boldsymbol x}
\newcommand{\mystateProjection}{\boldsymbol {\tilde x}}
\newcommand{\mystateDummy}{\boldsymbol y}

\newcommand{\mystateApprox}{\tilde \mystate}
\newcommand{\mystateInitial}{\mystate^0}
\newcommand{\mystateInitialArg}[1]{\mystateInitial\left(#1\right)}
\newcommand{\mystateRef}{\bar\mystate}
\newcommand{\velocity}{\boldsymbol f}
\newcommand{\velocityRed}{\velocity_r}
\newcommand{\velocityRedDis}{\boldsymbol g}
\newcommand{\func}{\boldsymbol h}
\newcommand{\velocityRedApprox}{\hat \velocityRed}
\newcommand{\nstate}{N}
\newcommand{\params}{\boldsymbol{\mu}}

\newcommand{\paramDomain}{\mathcal D}
\newcommand{\nparams}{p}
\newcommand{\paramsDummy}{\boldsymbol \nu}
\newcommand{\trialbasis}{\boldsymbol V}
\newcommand{\nstateRed}{n}
\newcommand{\ntrain}{N_\text{training}}
\newcommand{\nvalidation}{N_\text{validation}}
\newcommand{\mystateRed}{\hat{\mystate}}
\newcommand{\mystateRedDummy}{\hat\mystateDummy}

\newcommand{\nTimesteps}{N_t}

\newcommand{\timevar}{t}
\newcommand{\timevarDummy}{\tau}
\newcommand{\timeFinal}{T}
\newcommand{\timeDomain}{[0,\timeFinal]}

\newcommand{\RRplus}{\mathbb R_+}

\DeclareMathOperator*{\argmax}{argmax}
\DeclareMathOperator*{\argmin}{argmin}

\title{{Non-intrusive Nonlinear Model Reduction via Machine Learning Approximations to Low-dimensional Operators}}
\author{{\hspace{1mm}Zhe Bai}\\
	Computational Research\\
	Lawrence Berkeley National Lab\\
	Berkeley, CA 94720 \\
	\And
	{\hspace{1mm}Liqian Peng}\\
	Facebook AI Applied Research\\
	Facebook\\
	Menlo Park, CA 94025 \\
}
\date{}
\renewcommand{\shorttitle}{Non-intrusive Nonlinear Model Reduction via Machine Learning Approximations to Low-dimensional Operators}

\begin{document}
\maketitle
\begin{abstract}
Although projection-based reduced-order models (ROMs) for parameterized nonlinear dynamical systems have demonstrated exciting results across a range of applications, their broad adoption has been limited by their intrusivity: implementing such a reduced-order model typically requires significant modifications to the underlying simulation code. To address this, we propose a method that enables traditionally intrusive reduced-order models to be accurately approximated in a non-intrusive manner. Specifically, the approach approximates the low-dimensional operators associated with projection-based reduced-order models (ROMs) using modern machine-learning regression techniques. The only requirement of the simulation code is the ability to export the velocity given the state and parameters as this functionality is used to train the approximated low-dimensional operators. In addition to enabling nonintrusivity, we demonstrate that the approach also leads to very low computational complexity, achieving up to a three order of magnitude reduction in run time. We demonstrate the effectiveness of the proposed technique on two types of PDEs.
\end{abstract}

\keywords{model reduction, machine learning, low-dimensional operators, dynamical systems}

\section{Introduction}\label{sec1}
Modern computational architectures have enabled the detailed numerical simulation of incredibly complex physical and engineering systems at a vast range of scales in both space and time~\citep{Lee2013sc}.  
Even with improved high-performance computing, the iterative numerical solutions required for design and optimization may quickly become intractable; the computational demands for real-time feedback control are even more challenging~\citep{Brunton2015amr}. 
Fortunately, many high-dimensional dynamical systems, such as a discretized simulation of a fluid flow, are characterized by energetic coherent structures that evolve on an attractor or manifold with a lower dimensional intrinsic dimension~\citep{aubry1988dynamics,berkooz1993proper,holmes_lumley}.  
This observed low-dimensionality has formed the basis of reduced-order modeling, which now plays a central role in modern design, optimization, and control of complex systems. 
Despite the growing importance of model reduction, the challenges of nonlinearity, identifying an effective low-dimensional basis, and multiscale dynamics have limited its widespread use across the engineering and natural sciences~\citep{Brunton2018book}.  
A leading method for reduced-order modeling involves the Galerkin projection of known governing equations onto a mathematical or empirical basis, although this approach is intrusive and challenging to implement.  
The leading alternative involves black-box system identification, which is purely data-driven, but generally yields uninterpretable models.  
In this work, we investigate and compare several emerging techniques from machine learning, i.e. applied data-driven optimization, for non-intrusive reduced-order modeling.  

Intrusive model reduction methods, based on a working and decomposable numerical simulation of the governing equations, provide the most general and widely used set of techniques.  
Foremost in this arsenal is the Galerkin projection of the governing equations onto a low-dimensional linear subspace, usually spanned by orthogonal modes, such as Fourier, or data-driven modes from proper orthogonal decomposition (POD)~\citep{aubry1988dynamics,berkooz1993proper,holmes_lumley,Noack2003jfm}.  
These approaches benefit from a strong connection to the underlying physics, the ability to include constraints to enforce  preservation of the underlying dynamic structure, and adaptivity~\citep{Amsallem_2009,Carlberg2013jcp,Carlberg2015siamjsc,Carlberg2017jcp}.  
In addition, there are several extensions around hyper-reduction to improve efficiency for systems with parametric dependence or higher order nonlinearities, based on the empirical interpolation method (EIM)~\citep{Barrault2004crm} and the discrete empirical interpolation method (DEIM)~\citep{Chaturantabut2010siamjsc}. 
However, these models are limited to the dynamic terms of the governing equations and the linear projection basis may not be optimal for capturing the dynamics.
The main drawback of intrusive methods is that they are challenging to implement, requiring considerable human effort and knowledge to manipulate the governing equations and simulation code.  

In contrast to intrusive model reduction, data-driven techniques are becoming increasingly prevalent, catalyzed by the increasing availability of measurement data, the lack of governing equations for many modern systems, and emerging methods in machine learning and data-driven optimization.    
Collectively, these data-driven techniques form the field of system identification~\citep{ljung:book,Juang1994book,Nelles2013book}.  
Many techniques in system identification use regression to identify linear models, such as the eigensystem realization algorithm (ERA)~\citep{Juang1985jgcd} and dynamic mode decomposition (DMD)~\citep{Schmid2010jfm,Tu2014jcd,Kutz2016book,bai2020aiaa}; recently, both techniques have been connected to nonlinear systems via the Koopman operator~\citep{Mezic2005nd,Rowley2009jfm,Brunton2017natcomm}.  
Another classic technique in nonlinear system identification is the NARMAX algorithm~\citep{Kukreja2007nonlinear,Billings2013book,Semeraro2016arxiv}.  
Recently, powerful techniques in machine learning are re-defining what is possible in system identification, leveraging increasingly large datasets and more powerful optimization and regression techniques.  
Deep learning, based on multi-layer neural networks, is increasingly used to model fluid dynamics and obtain closure models~\citep{Milano2002jcp,Ling2016jfm,Zhang2015aiaa,Singh2017aiaaj,Duraisamy2018arfm,Wang2017prf,Kutz2017jfm}.  
More generally, machine learning is having a considerable impact in the modeling of dynamical systems and physics~\citep{Bongard2007pnas,Schmidt2009science,Brunton2016pnas,Raissi2018jcp}, for example relevant work in cluster-based reduced order models~\citep{Kaiser2014jfm}, long-short time memory networks (LSTMs)~\citep{wan2018data,vlachas2018data}, and Gaussian process regression~\citep{Raissi2017arxiva,Wan2017physicad}. 
Of particular interest are techniques based on the principle of \emph{parsimony} that seek the simplest models, with the fewest terms necessary to describe the observed dynamics~\citep{Bongard2007pnas,Schmidt2009science,Brunton2016pnas,Rudy2017sciadv,Loiseau2017jfm,Loiseau2018jfm,Schaeffer2017prsa}.    
For example, the sparse identification of nonlinear dynamics (SINDy) algorithm~\citep{Brunton2016pnas} uses sparse regression to identify the few active terms in a differential equation.  
The resulting models balance model complexity with descriptive power, avoiding overfitting and remaining both interpretable and generalizable.  

Data-driven techniques in system identification and machine learning are currently being employed for advanced, non-intrusive, and reduced-order modeling, removing the need for full-order model equations or a modifiable numerical code.  
Loiseau and Brunton~\citep{Loiseau2017jfm} extended the SINDy modeling framework to enforce known constraints, such as energy conserving quadratic nonlinearities in incompressible flows.  
This so-called \emph{Galerkin regression} also enables higher order nonlinearities than are possible with standard Galerkin projection, providing effective closure for the dynamics of truncated modes.  
Swischuk et al.~\citep{SWISCHUK2018} develop parametric models between input parameters and POD coefficients for physics and engineering problems, comparing many leading methods in machine learning. 
Peherstorfer and Willcox~\citep{PEHERSTORFER2016196} develop time-dependent reduced order models for partial differential equations, by non-intrusively inferring the reduced-order model operator output as a function of initial conditions, inputs, trajectories of states, and full-order model outputs, without requiring access to the full-order model. 
They prove that for dynamics that are linear in the state or are characterized by low-order polynomial nonlinearities, the ROM converges to the projection of the FOM onto a low-order subspace.
Carlberg et al.~\citep{Carlberg2019JCP} propose dimensionality reduction on autoencoders to learn dynamics for recovering missing CFD data.
Other non-intrusive ROMs have been developed based on neural networks~\citep{Wang_2016,Kani:2017aa,Xie:2018aa,HESTHAVEN201855,Wang:255708}, radial basis functions~\citep{Xiao_2015}, and kernel methods~\citep{Vaerenbergh:2010aa,wirtz2015ijnme}.  
Hybrid methods are also promising, for example by modeling the error of a Galerkin model to produce closures~\citep{doi:10.1137/17M1145136}.  

Despite the considerable recent progress in leveraging machine learning for non-intrusive reduced-order modeling, current approaches still have a number of shortcomings.  
First, many methods are limited to low-order polynomial nonlinearities.  
All of the above approaches learn the low-dimensional operators from full-system trajectories, which is expensive and limits the predictive accuracy, as the ROM trajectory is likely to quickly deviate from the full-system trajectory.  
Current methods also have limited consideration for stability and the interplay between regression methods and time integration has not been explored.  
Finally, current approaches do not provide a framework for model selection, as each method will likely be best suited to a different problem class.  

In this work, we continue to develop and explore machine learning techniques for non-intrusive model reduction, addressing many of the shortcomings described above.
In particular, we develop and compare numerous leading techniques in machine learning to produce accurate and efficient reduced order models.  
We focus primarily on the class of Galerkin projection models, although these approaches may be extended to more general models. 
To study the dynamical system in the latent space, we map from the low-dimensional state at the current time instance to the next, in explicit or implicit time integration schemes. 
We investigate a variety of regression models for the mapping, including k-nearest neighbors~\citep{altman1992knn}, support vector regression~\citep{smola2004svm}, random forest~\citep{breiman2001random}, boosted trees~\citep{hastie2009elements}, the vectorial kernel orthogonal greedy algorithm (VKOGA)~\citep{wirtz2013special, wirtz2015ijnme} and SINDy~\citep{Brunton2016pnas}.
In the following, we explicitly consider stability and explore the connection between various regression methods and their suitability with different time integration schemes; for example, non-differentiable machine learning models are only amenable to explicit time integration.  
We also explore the modeling procedures of varying complexities on two examples, the 1D Burgers' equation and the 2D convection-diffusion equation using both explicit and implicit time integrators.

\section{Problem formulation}\label{sec2}

\subsection{General nonlinear dynamical systems} \label{sec2.1}

 This work considers parameterized nonlinear dynamical systems characterized
 by the system of nonlinear ODEs
 \begin{eqnarray} \label{eq:ODEFOM}
\dot \mystate &= \velocity(\mystate;\timevar,\params),\qquad
\mystate(0;\params) &=\mystateInitial(\params),
  \end{eqnarray} 
	where
	$\timevar\in\timeDomain$ denotes time with $\timeFinal\in\RRplus{}$ representing
	the final time,
	$\mystate\in\RR{\nstate}$ denotes the state,
	$\params\in\paramDomain\subseteq\RR{\nparams}$ denotes the system
	parameters, $\mystateInitial:\paramDomain\rightarrow\RR{\nstate}$ is the
	parameterized initial condition, and
	$\velocity:\RR{\nstate}\times\RRplus{}\times\paramDomain\rightarrow\RR{\nstate}$
	denotes the velocity. 

 \subsection{Galerkin projection} \label{sec2.2}
We assume that we  we have low-dimensional trial-basis
	matrix $\trialbasis\in\RRstar{\nstate\times \nstateRed}$ (with
	$\nstateRed\ll\nstate$) computed, e.g., via proper orthogonal decomposition
	(POD), such that the solution can be approximated as
	$\mystate(t,\params)\approx\mystateApprox(t,\params) = \mystateRef(\params) +
	\trialbasis \mystateRed(t,\params) \approx \mystate$ with
	$\mystateRed\in\RR{\nstateRed}$ denoting the reduced state. 
	Then, the Galerkin ROM can be expressed as
 \begin{align} \label{eq:Galerkin}
\dot \mystateRed &= \trialbasis^T\velocity(\trialbasis \mystateRed,\timevar;\params)\\
\mystateRed(0) &=\trialbasis^T \mystateInitialArg{\params}.
  \end{align} 
Critically, note that the ROM is defined by the low-dimensional mapping
\begin{align}
\velocityRed:&(\mystateRedDummy,\timevarDummy;\paramsDummy)\mapsto
\trialbasis^T\velocity(\trialbasis\mystateRedDummy,\timevarDummy;\paramsDummy)\\
:&\RR{\nstateRed}\times\RRplus{}\times\RR{\nparams}\rightarrow \RR{\nstateRed}.
\end{align}
The reduced velocity $\velocityRed$ is thus simply a function that maps the
reduced state and inputs to a low-dimensional vector.  
Thus, we can rewrite Equation~(\ref{eq:Galerkin}) as
\begin{equation} \label{eq:Galerkin_compact}
 \dot \mystateRed = \velocityRedApprox (\mystateRed,\timevar;\params)
\end{equation}

However, this approach is
intrusive to implement in computational-mechanics codes,
as it requires querying the full-order model code to compute
$\velocity(\trialbasis\mystateRed,\timevar;\params)$
for every instance of $\mystateRed$ and $\params$ encountered during the ROM
	simulation; without hyper-reduction (e.g., DEIM or gappy POD), it is
also computationally expensive, as computing $\trialbasis^T\velocity$ given
$\velocity$ incurs $\mathcal O(\nstate\nstateRed)$ flops. Even if
hyper-reduction is present, however, the approach remains intrusive, as
sampled entries of the velocity must be computed for every instance of $\mystateRed$ and $\params$ encountered during the ROM
	simulation.

\section{Regression-based reduced operator approximation}\label{sec3} 

This section outlines learning the low dimensional operator through regression methods. We commence describing the general formulation of the framework in Section~\ref{sec3.1}, and then Section~\ref{sec3.2} discusses the proposed regression models and their computational complexity. In Section~\ref{sec3.3}, we derive the boundedness and stability of of the approximated discrete-time dynamical system.   

\subsection{Mathematical formulation}\label{sec3.1}

The objective is to develop a data-driven, non-intrusive approximation to the reduced velocity with reduced-order models of different dynamical systems.
$\velocityRed$. In particular, we aim to devise an approximate mapping
\begin{align}
\velocityRedApprox
:&\RR{\nstateRed}\times\RRplus{}\times \RR{\nparams}\rightarrow \RR{\nstateRed}.
\end{align}
such that
\begin{align}
\velocityRedApprox(\mystateRed,\timevar;\params)\approx\velocityRed(\mystateRed,\timevar;\params),\quad\forall
(\mystateRed,\timevar;\params)\in\RR{\nstateRed}\times\RRplus{}\times\RR{\nparams}.
\end{align}
As the domain and codomain of $\velocityRedApprox$ exhibit no specialized
structure, we can consider general high-dimensional regression techniques
(e.g., kernel regression, random-forest regression).

Similar to  Equation~(\ref{eq:Galerkin_compact}),   the reduced order model is given by
\begin{equation} \label{rom}
 \dot \mystateRed = \velocityRedApprox (\mystateRed,\timevar;\params),
\end{equation}
with the initial condition 
\begin{equation} \label{rom_init}
 \mystateRed(0) = \trialbasis^T \mystateInitialArg{\params}.
\end{equation}

We assume that the sequences of reduced velocity $\velocityRed$ can be studied from the Markovian discrete-time dynamical system

\begin{equation} \label{rom_dis}
\mystateRed^{j+1} = \velocityRedDis(\mystateRed^{j}, \params), \quad j = 0,\cdots, \nTimesteps,
\end{equation}
for discrete-time velocity $\velocityRedDis: \RR{\nstateRed}\times \RR{\nparams} \rightarrow \RR{\nstateRed}$. If such a reduced velocity exists and we can compute the operator, the whole time sequence of the reduced states can then be estimated from Equation~(\ref{rom_dis}) by specifying the initial (reduced) state. By considering the mapping between the current reduced state and the next, we cast the problem into a set of regression models.

\subsection{Surrogate ROM}\label{sec3.2}

In this work, we collect the reduced states from the Galerkin projection as in Section~\ref{sec2.2} and approximate the operator $\velocityRedDis$ by regressing each component of the reduced velocity. To be specific, we construct individual regression $\hat{g}_i \approx g_i$ for $i = 1, \cdots, \nstateRed$, the $i$th equation of $\velocityRedDis$ is:
\begin{equation} \label{rom_dis_i}
\hat{x}_i^{j+1} = \hat{g}_i(\mystateRed^{j}, \params), \quad j = 0,\cdots, \nTimesteps, \quad i = 1,\cdots,\nstateRed,
\end{equation}
where $\mystateRed = [x_1 \cdots x_n]$ and $\velocityRedDis = [g_1 \cdots g_n]$. Equation~(\ref{rom_dis_i}) shows that we can map the reduced state at time step $\mystateRed^{n}$ to each component of the state at the next time step $\hat{x}_i^{n+1}$ for $j = 0,\cdots, \nTimesteps - 1$. 

In the offline stage, we obtain training data from the existing Galerkin projection model. By querying both features ($\mystateRed^{j}$) and responses ($\hat{x}_i^{j+1}$) at every time instance, we generate a data set $\mathcal{T}_{train,i} = {(\mystateRed^{j}, x_i^{j+1})}$ for model training and cross validation.

We consider a variety of conventional statistical models, including support vector regression (SVR) with kernel functions (Gaussian, polynomials)~\citep{smola2004svm}, tree-based methods (boosted decision trees~\citep{hastie2009elements}, random forest~\citep{breiman2001random}) and $k$-nearest neighbors~\citep{altman1992knn}, for regressing the feature-response pairs. Three types of kernel functions are explored in SVR. Specifically, we refer to the SVR model using the 2$^{nd}$, 3$^{rd}$ order polynomial kernel as SVR2, SVR3 respectively. When the Gaussian radial basis function (RBF) kernel is used, we refer to the model as SVRrbf. In addition, we investigate the vectorial kernel orthogonal greedy algorithm (VKOGA) and Sparse identification of nonlinear dynamics (SINDy) as advanced regression models. More details of VKOGA and SINDy can be found in Appendix~\ref{app1}. We compare the computational complexity of all the proposed regression models in Appendix~\ref{app1}.
For all the candidate models, we employ cross validation to optimize the hyperparameters for model selection, aiming for a balanced bias and variance. We also explore the training and validation error as varying the number of samples in Appendix~\ref{app2}, and we select a fixed sample size for performance comparison between models.

For time integration along the trajectory of the dynamical system, we investigate the appropriate time step for each regression model. We implement the $4$th-order Runge-Kutta as the explicit method, as well as Newton-Raphson and fixed-point iteration both in backward Euler as the implicit methods. We report the numerical results, with respect to the Galerkin reduced-order model (ROM) and the full-order model (FOM) in a sequence of time in Section~\ref{Section4}.

\subsection{Error analysis}\label{sec3.3}
Assuming the considered regression model generates bounded in reduced space, we examine the boundedness of the surrogate FOM on the time evolution of the states along the trajectory. 	 
Let $J_t=[0, T]$ denote the time domain, $J_\mu \subset \RR{\nparams}$ be a compact space of $\params$,  $\mystate: J_t \to \mathbb{R}^n$ denote the state variable,    and $\velocity(\mystate,\timevar;\params)$ be the vector field. 
Let $ \velocityRed(\mystateRedDummy,\timevar;\params) = \velocity(\trialbasis \mystateRedDummy,\timevar;\params)$ represent the Galerkin reduced vector field. 
Let $\mystateRed (\timevar)$ denote an approximate reduced vector field constructed by a machine learning method and   $\velocityRedApprox(\mystateRed,\timevar;\params)$ be the corresponding vector field.   
 
The error of the reduced model  can be defined as $\error(\timevar) :=\mystateRed (\timevar) -\mystate (\timevar)$. Let $P = \trialbasis \trialbasis^*$. Let $\erroro(\timevar):=(I_n -P) \error(\timevar)$, which denotes the error component orthogonal to  range($\trialbasis$), and $\errori(\timevar):=P \error(\timevar)$, which denotes the component of error parallel to range($\trialbasis$). Let $\mystateProjection := P \mystate$ denote the direct projection of $\mystate$.
 Thus,  we have
\begin{equation}
\erroro(\timevar)=\mystateProjection(\timevar)- \mystate(\timevar).
\end{equation}
However, since the system is evolutionary with time, further approximations of the projection-based reduced model
result in an additional error $\errori(\timevar)$, and  we have
\begin{equation}
\errori(\timevar)=\trialbasis \mystateRed (\timevar)-\mystateProjection(\timevar).
\end{equation} 
Although $\errori(\timevar)$ and $\erroro(\timevar)$ are orthogonal to each other, they are not independent.

\textbf{lemma} \label{bound} Consider the dynamical system as Equation (\ref{eq:ODEFOM}) over the interval $\timevar \in J$ and $\params \in J_\mu$.  Suppose $\velocity(\mystate,\timevar;\params)$ is a uniformly Lipschitz function of $\mystate$ with constant
$K$ and a continuous function of $\timevar$  for all $(\mystate,\timevar, \params) \in  B_b(\mystate_0) \times  J_t  \times J_\mu$.
  Suppose $\velocityRedApprox(\mystateRed,\timevar;\params)$ is a uniformly Lipschitz function of  $\mystateRed$ and a continuous function of $\timevar$ for all $(\mystate,\timevar;\params) \in  B_b(x_0) \times  J_t  \times J_\mu$. Suppose the data-driven approximation satisfies $\| \velocityRedApprox(\mystateRed,\timevar;\params)  - \trialbasis^* \velocity (\trialbasis \mystateRed,\timevar;\params)  \| \le C$ for all  $(\mystateRed, \timevar, \params) \in  B_b(\mystateRed(\times_0)) \times J_t \times J_\mu$. Then the error $\error(\timevar)=\mystateRed (\timevar)-\mystate(\timevar)$ in the infinity norm for the interval $J_t$ is bounded by
\begin{equation}\label{eeo}
\left\| \error \right\|_\infty \le e^{KT}  \left\| \erroro \right\|_\infty +e^{KT} \left\| {\errori(0)} \right\|    + \frac{C}{K}(e^{KT}-1) 
\end{equation}

\textbf{proof:} For notation simplification, we fix $\params$ and do not explicitly denote it in vector fields. 

Substituting Equation~(\ref{eq:ODEFOM}) and (\ref{rom}) into the differentiation of $\erroro(\timevar)+ \errori(\timevar) = \trialbasis \mystateRed(\timevar) - \mystate(\timevar)$ yields
\begin{equation}\label{ee}
\dot \erroro + \dot \errori = \trialbasis \velocityRedApprox(\mystateRed,\timevar)  - \velocity(\mystate, \timevar) .
\end{equation}
Left multiplying (\ref{ee})  by $P$ and recognizing that $P \trialbasis = \trialbasis$ gives
\begin{equation*}
\begin{aligned}
 \dot \errori(\timevar) &= \trialbasis \velocityRedApprox(\mystateRed,\timevar)  - P \velocity(\mystate, \timevar) \\
&= \trialbasis \velocityRedApprox(\mystateRed,\timevar) -P\velocity(\trialbasis \mystateRed, \timevar) + P\velocity (\trialbasis \mystateRed, \timevar)  - P\velocity( \mystate+\erroro, \timevar)+P\velocity(\mystate+\erroro, \timevar)-P \velocity(\mystate, \timevar) \\
&=\trialbasis (\velocityRedApprox(\mystateRed, \timevar) - \trialbasis^* \velocity(\trialbasis \mystateRed, \timevar)) +P(\velocity(\mystate+\erroro+\errori, \timevar)-\velocity(\mystate+\erroro, \timevar))\\
& \quad  +P(\velocity(\mystate+\erroro, \timevar)- \velocity(\mystate, \timevar) )
\end{aligned}
\end{equation*}

 Using this equation by expanding  $ \|\errori(\mystate + h)  \|$ and  applying triangular inequality yields
\begin{equation*}
\begin{aligned}
 \| \errori(\timevar + h) \| &= \| \errori(\timevar) + h \dot \errori(\timevar) \| + \mathcal{O}({h^2}) \\
&  \le \| \errori(\timevar) \| +  \| h \trialbasis (\velocityRedApprox(\mystateRed, \timevar) - \trialbasis^* \velocity(\trialbasis \mystateRed, \timevar))  \|  +   \| h P (\velocity(\mystate + \erroro + \errori, \timevar) \\ & \quad - \velocity(\mystate + \erroro, \timevar)) \|   + \| h P (\velocity(\mystate + \erroro, \timevar) - \velocity(\mystate, \timevar) ) \| + \mathcal{O}(h^2)\\
&  \le \| \errori (\timevar) \| + h \| \trialbasis \|\cdot \| \velocityRedApprox(t,\mystateRed) - \trialbasis^* \velocity(\trialbasis \mystateRed, \timevar)  \|  +  h\| P  \| \cdot  \| \velocity(\mystate + \erroro + \errori, \timevar)  \\
& \quad - \velocity(\mystate + \erroro, \timevar) \|   + h\| P  \| \cdot  \|  \velocity(\mystate +\erroro, \timevar) - \velocity(\mystate, \timevar)  \| + \mathcal{O}(h^2).
\end{aligned}
\end{equation*}
Using $\|\trialbasis\|=\|P\|=1$ and $\| \velocityRedApprox(\timevar, \mystateRed) - \trialbasis^* \velocity(t, \trialbasis \mystateRed)  \| \le C$, the last inequality gives
\begin{equation*}
\begin{aligned}
 \| \errori(\timevar + h) \| & \le \| \errori (\timevar) \| + h C  +  h \| \velocity(\mystate + \erroro + \errori, \timevar)    - \velocity(\mystate, + \erroro, \timevar) \|   \\
 &\quad +  h  \|  \velocity(\mystate +\erroro, \timevar) - \velocity(\mystate, \timevar)  \| + \mathcal{O}(h^2).
\end{aligned}
\end{equation*}
Rearranging this inequality and applying the  Lipschitz conditions gives
\begin{equation*}
\frac{\| \errori(\timevar + h) \| - \|\errori(\timevar) \|}{h} \le K \| \errori (\timevar) \| + K \| \erroro (\timevar) \| + C+ \mathcal{O}(h).
\end{equation*}
Since $\mathcal{O}(h)$ can be uniformly bounded independent of $\errori(\timevar)$, using the mean value theorem and letting $h \to 0$ give
\begin{equation*}
\frac{d}{dt} \|\errori (\timevar)\| \le K\| \errori(\timevar) \| + K\| \erroro(\timevar) \|+C.
\end{equation*}
Rewriting the above inequality into integral form, $\left\| \errori(\timevar) \right\| \le \alpha (\timevar) + K \int_0^ \timevar \| \errori(\tau ) \| d\tau$, where $\alpha (\timevar): = \left\| \errori(0) \right\| + K\int_0^ \timevar {\| \erroro(\tau ) \| d \tau }+C \timevar$, and using Gronwall's
lemma, we obtain
\begin{equation*}
\left\| {{\errori}(\timevar)} \right\| \le \alpha (\timevar) + \int_0^ \timevar {\alpha (s)K\exp \left( {\int_s^ \timevar {Kd\tau } } \right) ds}.
\end{equation*}
By definition, ${\left\| \erroro \right\|_ \infty} \ge \left\|
\erroro(\timevar) \right\|$ for any $\timevar \in J_t$. It follows that $\alpha (\timevar) \le \left\| {{\errori}(0)} \right\| + K \timevar\left\| \erroro \right\|_\infty+C \timevar$. Simplifying the integral of the right-hand side of the above inequality gives
\begin{equation*}
 \| \errori(\timevar) \|  \le (e^{K \timevar}-1) \left(\| \erroro \|_\infty+ \frac{C}{K} \right)+  e^{K \timevar} \|\errori(0) \|,
\end{equation*}
for any $\timevar \in J_t$. If follows that
\begin{equation*}
 \| \errori \|_\infty  \le (e^{KT}-1) \left(\| \erroro \|_\infty+ \frac{C}{K} \right)+  e^{KT} \|\errori(0) \|,
\end{equation*}
Combining the above inequality with $\left\| \error \right\|_\infty
\le \left\| {\errori } \right\|_\infty + \left\| {\erroro }
\right\|_\infty$, one can obtain Equation~(\ref{eeo}).

\textbf{Remark:} The above lemma provides a bound for $\|\errori(\timevar)\|$ in terms of $\|\erroro\|_\infty$ and $\|\errori(0)\|$. We have $\left\|\errori(0)\right \|=0$ when the
initial condition of the reduced model is given by $\mystateRed(0)=\trialbasis^* \mystate_0$ for Equation~(\ref{rom}). In this situation, Equation~(\ref{eeo}) becomes ${\left\| \error\right\|_\infty} \le e^{KT} {\left\| {{\erroro}} \right\|_\infty}+  \frac{C}{K}(e^{KT}-1) $. 

\section{Numerical experiments}\label{Section4}
To assess the proposed non-intrusive ROM, we consider two parameterized PDEs: (i) 1D inviscid Burgers' equation, and (ii) 2D convection--diffusion equation. We implement explicit integrator, including 4th-order Runge-Kutta solvers, and Newton-Raphson, fixed-point iteration in backward Euler as the implicit methods.
\subsection{1D Burgers' equation}
The experiment first employs a 1D parameterized inviscid Burgers' equation. The input parameters $\params=(a,b)$ in the space $[1.5, 2]\times[0.02, 0.025]$.  In the current setup, the parameters for online test are fixed to be $\params=(1.8, 0.0232)$. In the FOM, the problem is often solved using a conservative finite-volume formulation and Backward Euler in time. The 1D domain is discretized using a grid with 501 nodes, corresponding to $x=i\times(100/500), i=0,\cdots,500$. The solution $\solution(x,\timevar)$ is computed in the time interval $\timevar\in [0,25]$ using different time step sizes considering the convergence in each time integrator.
\begin{subequations}\label{eq:Burgers'}
\begin{align} 
\tag{\ref{eq:Burgers'}}\frac{\partial \solution(x,\timevar)}{\partial \timevar} + \frac{1}{2} \frac{\partial (\solution^2(x,\timevar))}{\partial x} &= 0.02e^{bx},\\
\label{eq:Burgers'BC}\solution(0,\timevar) &= a, \forall \timevar>0,\\
\label{eq:Burgers'IC}\solution(x,0) &= 1, \forall x \in [0,100]
\end{align}
\end{subequations}
The solution $\solution(\mystate, \timevar)$ is computed in the time interval $\timevar \in [0,25]$ using a uniform computational time-step size $\Delta\timevar$. 

 \subsubsection{Data collection}
 	
\begin{figure}
    \centering
	\begin{overpic}[width = 0.95\textwidth]{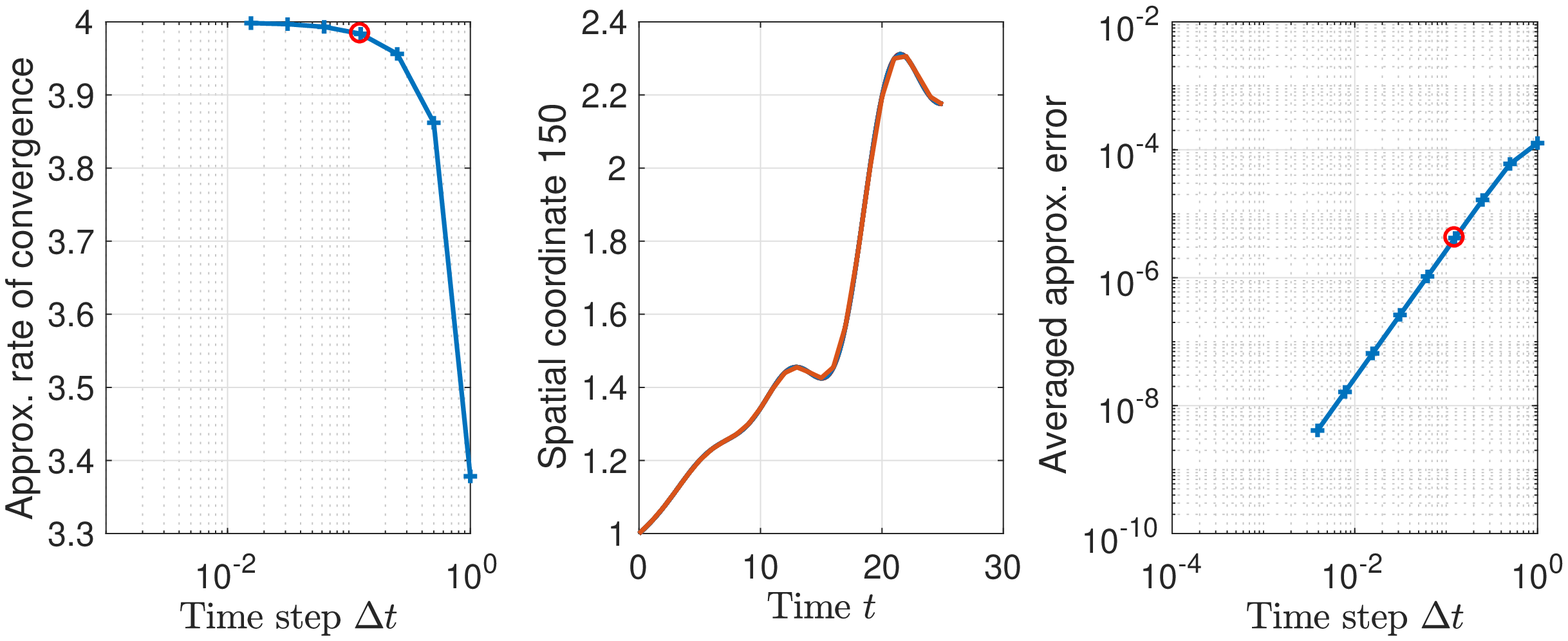}
		\put(3,32){(a)}
	\end{overpic}\\
	\begin{overpic}[width = 0.95\textwidth]{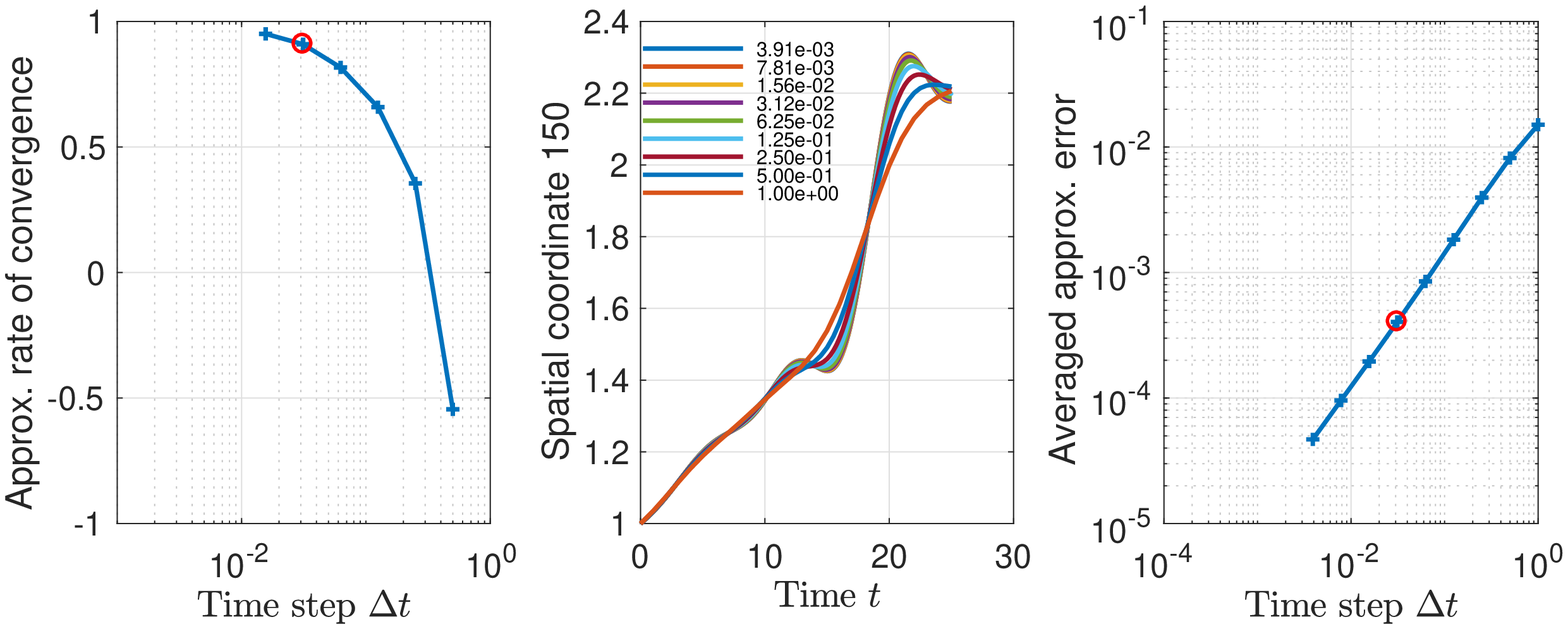}
		\put(3,32){(b)}
        \put(55,31){\vector(1,-1){4.5}}
        \put(57,24){$\Delta t$}
	\end{overpic}
	\caption{Timestep verification for 1D inviscid Burgers' equation: (a) Runge-Kutta: we select a time step of $1.25e{\text -}1$, as it leads to an approximated ($<1\%$ error) convergence rate of $4$  and an averaged approximated error of $6e{\text -}5$ for the selected state. (b) backward-Euler integrator: we select a time step of $3.12e{\text -}2$, as it leads to an approximated convergence rate of 1 and an approximated error of $4e{\text -}4$ for the selected state.}
	\label{FigTimestep}
\end{figure}
We investigate time step verification on choosing an appropriate $\Delta t$ of the time integrator for the problem. We collect the solutions under an increasing number of time steps $\nTimesteps = [25, 50, 100, 200, 400,  800, 1600, 3200, 6400]$ using both Runge-Kutta as well as backward Euler integrator. Throughout the paper, we select the time step at $99\%$ of the asymptotic rate of convergence. 
The verification results in Figure~\ref{FigTimestep} show that $\nTimesteps = 200$ is a reasonable number of time steps to use for the 4th-order Runge-Kutta and $nT=800$ for backward Euler method.
During the offline stage, we run four full simulations corresponding to the corner parameters of the space $[1.5, 2]\times[0.02, 0.025]$. Then, we sample the training data from Latin-hypercube, . In the sampling, $\ntrain$ and  $\nvalidation$ instances of the state, time and parameters are generated following the criterion that the minimum distances between the data points are maximized.  For this study, the default size of the training set is $\ntrain = 1000$ and the default size of the validation set is $\nvalidation = 500$. The reduced vector field $\velocityRed$ is computed for each input pairs $(\mystateRed,\timevar;\params)$. Note that both the training and validation stage only involves pure machine learning. Then in the test stage, we evaluate the proposed ROM. The parameters are fixed to be $\mu=(1.8,0.0232)$ for testing purpose.
 
  \subsubsection{Model validation} \label{Section4.1.3}
  	
	We use SVR with kernel functions (2$^{nd}$, 3$^{rd}$ order polynomials and radial basis function), kNN, Random Forest, Boosting, VKOGA, and SINDy as regression models to approximate reduced velocity. 
  In particular, for each regression method, we change the model hyperparameters and plot the relative training error and validation error. The relative error is defined by 
  
  \begin{equation}\label{err}
	err = \frac{\| \velocityRedApprox(\mystateRed,\timevar;\params) - \velocityRed(\mystateRed,\timevar;\params)\|}{\| \velocityRed(\mystateRed,\timevar;\params)\|}. 
  \end{equation}
	
We then plot the learning curve of each regression method and compare the performance of each model on training and validation data over a varying number of training instances in Appendix~\ref{app2}. By properly choosing the hyperparameters and the number of training instances, our regression models can effectively balance bias and variance.

\subsubsection{Simulation of the surrogate ROM} \label{Section4.1.4}
We can now solve the problem using the surrogate model along the trajectory of the dynamical system. After applying time integration to the regression-based ROM, we compute the relative error of the proposed models as a function of time. We investigate both Newton-Raphson, fixed-point iteration in backward Euler and 4th-order Runge-Kutta in explicit methods.
Let $\mystate(\timevar)$, $\mystateGalerkin(\timevar)$ and $\mystateRed(\timevar)$ denote the solution of the FOM, the Galerkin ROM, and non-intrusive ROMs respectively. We define the relative error with respect to FOM $\errorFOM$ and Galerkin ROM $\errorROM$ as 

\begin{equation}
\errorFOM = \frac{\| (\trialbasis \mystateRed(\timevar) - \mystate (\timevar))\|}{\| \mystate (\timevar)\|},
\end{equation}
\begin{equation}
\errorROM = \frac{\| ( \mystateRed(\timevar) - \mystateGalerkin (\timevar))\|}{\| \mystateGalerkin (\timevar)\|},
\end{equation}
The corresponding averaged relative error over the entire time domain $\errorFOMave$ and $\errorROMave$ can be computed as
\begin{equation}
\errorFOMave = \frac{1}{T}\int_{t=0}^T \errorFOM dt,
\end{equation}
\begin{equation}
\errorROMave = \frac{1}{T} \int_{t=0}^T \errorROM dt.
\end{equation}
Let $t_{FOM}$, $t_{ROM}$, and $\tau$ denote the running time of FOM,  Galerkin ROM, and non-intrusive ROM respectively. Define  the relative running time with respect to the FOM and the Galerkin ROM:

 \begin{equation}
\tauFOM = \frac{\tau} {t_\textit{FOM}},
\end{equation}
 and 
  \begin{equation}
  \tauROM = \frac{\tau}{t_\textit{ROM}}.
 \end{equation}

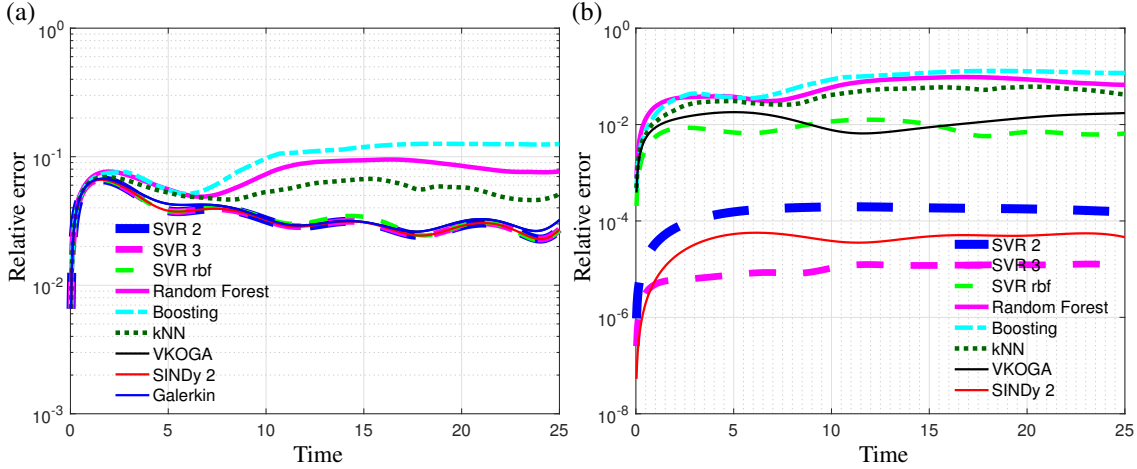
\begin{figure}
	\centering
	\begin{overpic}[width = 0.45\textwidth]{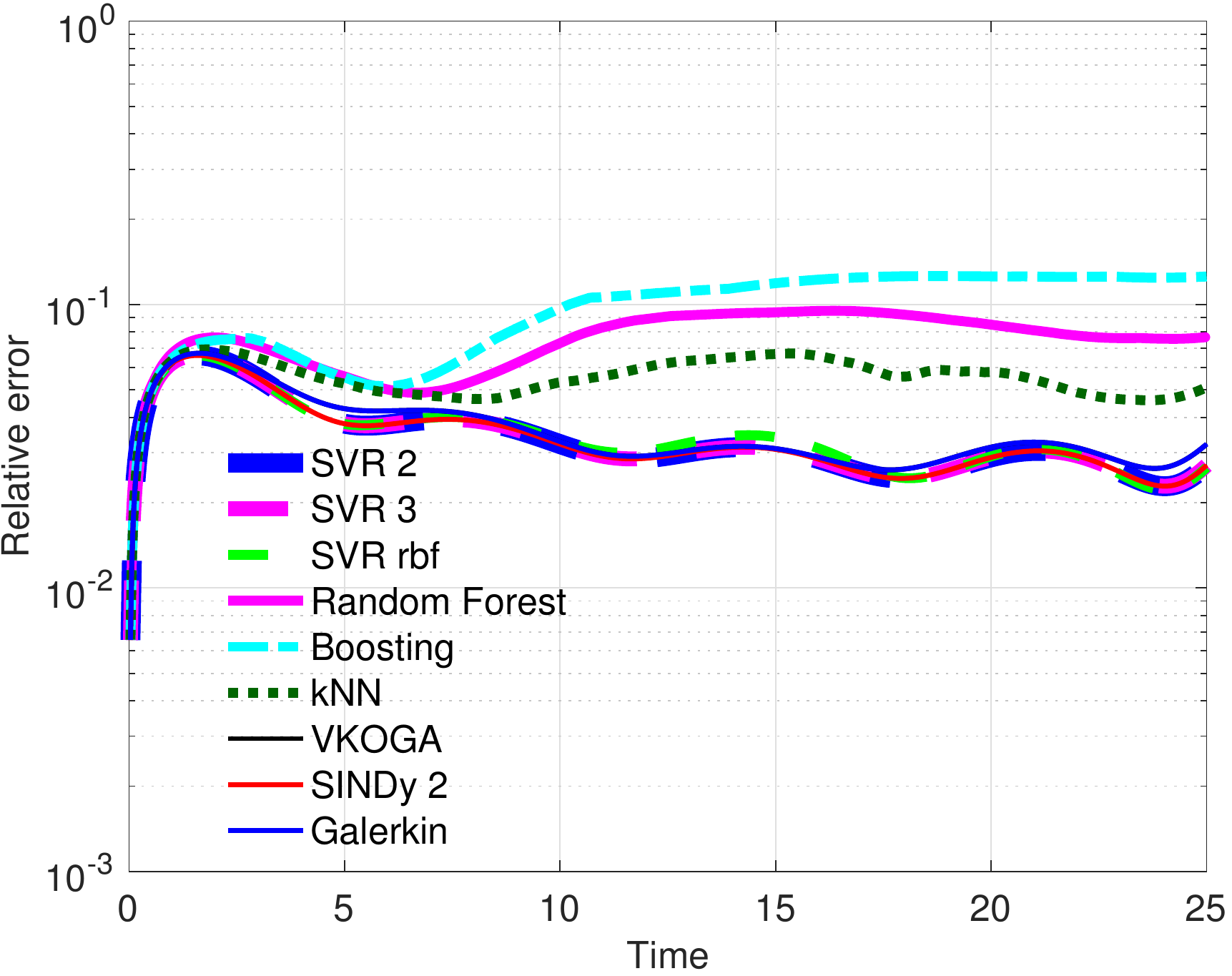}
		\put(-1,79){(a)}
		\put(49,-0.1){\colorbox{white}{\small{Time}}}
		\put(-2,32){\colorbox{white}{\rotatebox{90}{\small{Relative error}}}}
	\end{overpic}
	\begin{overpic}[width = 0.45\textwidth]{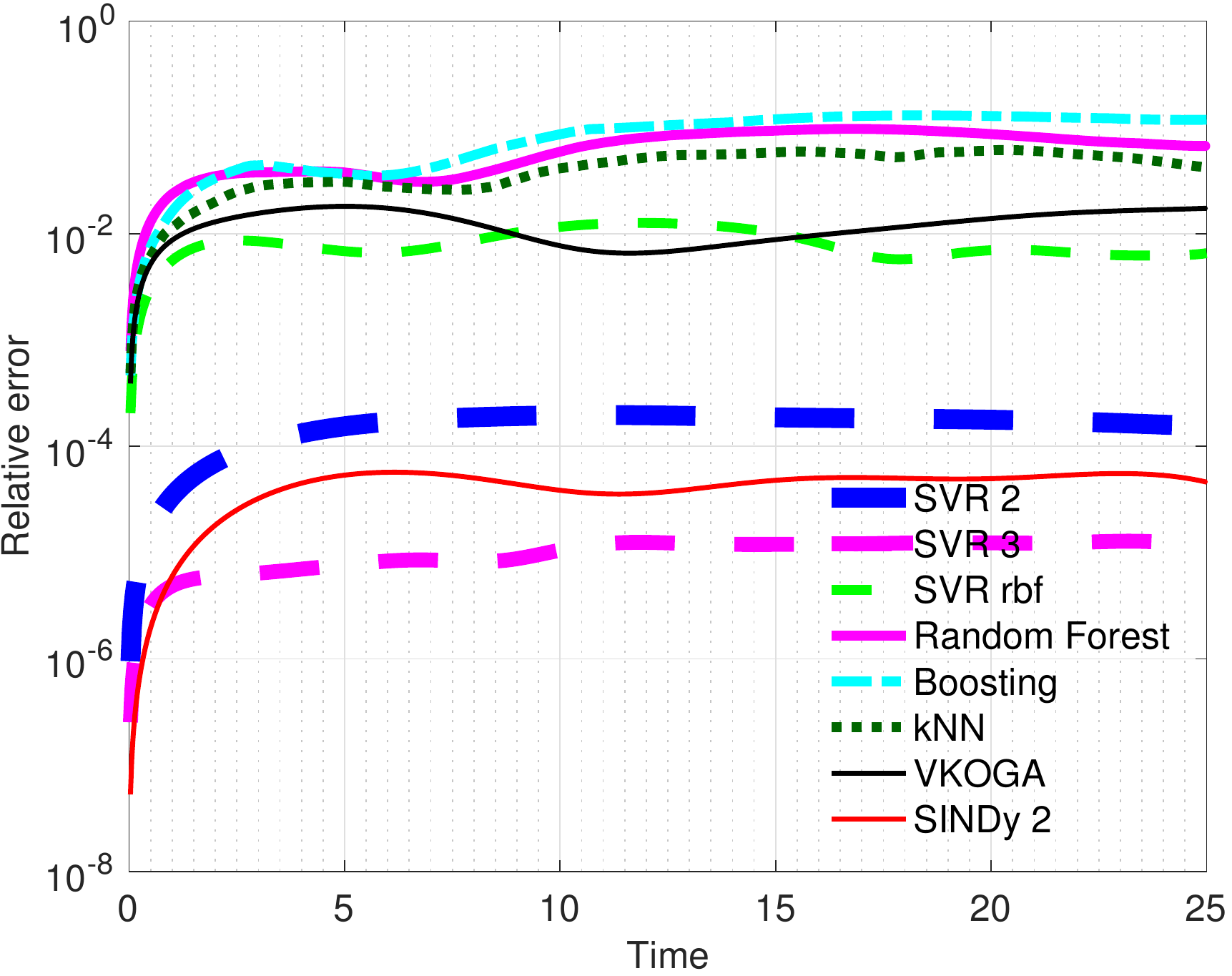}
		\put(-1,79){(b)}
		\put(49,-0.1){\colorbox{white}{\small{Time}}}
		\put(-2,32){\colorbox{white}{\rotatebox{90}{\small{Relative error}}}}
	\end{overpic}
	\caption{Backward Euler for 1D inviscid Burgers' equation: time evolution of relative error: (a) $\errorFOM$ in FOM; and  (b) $\errorROM$ in ROM.}
	\label{BEtime}
 \end{figure}

\begin{figure}
	\centering
	\vspace{.3in}
	\begin{overpic}[width = 0.45\textwidth]{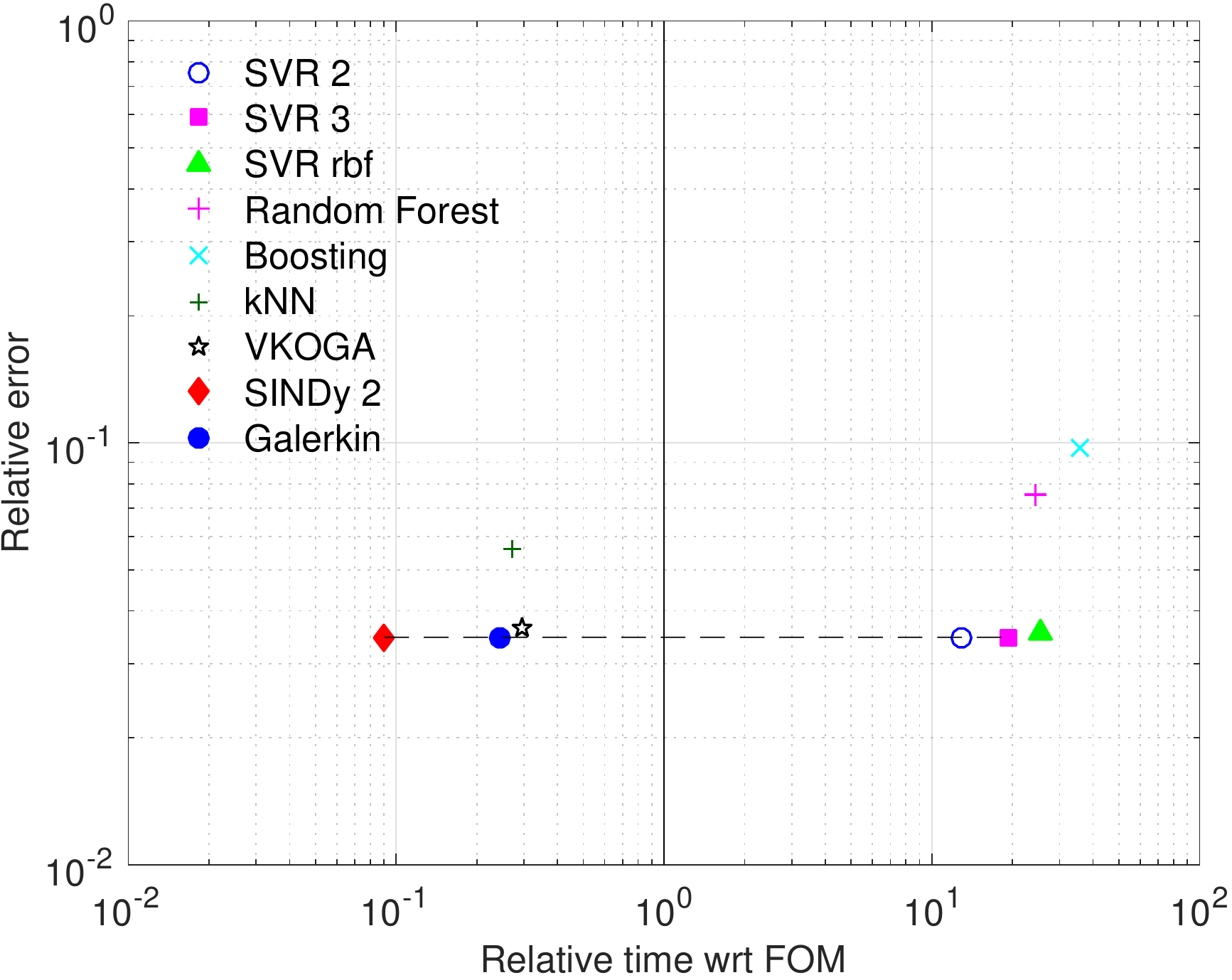}
		\put(-1,79){(a)}
		\put(-2,32){\colorbox{white}{\rotatebox{90}{\small{Relative error}}}}
		\put(32,-0.1){\colorbox{white}{\small{Relative time w.r.t. FOM}}}
	\end{overpic}
	\begin{overpic}[width = 0.47\textwidth]{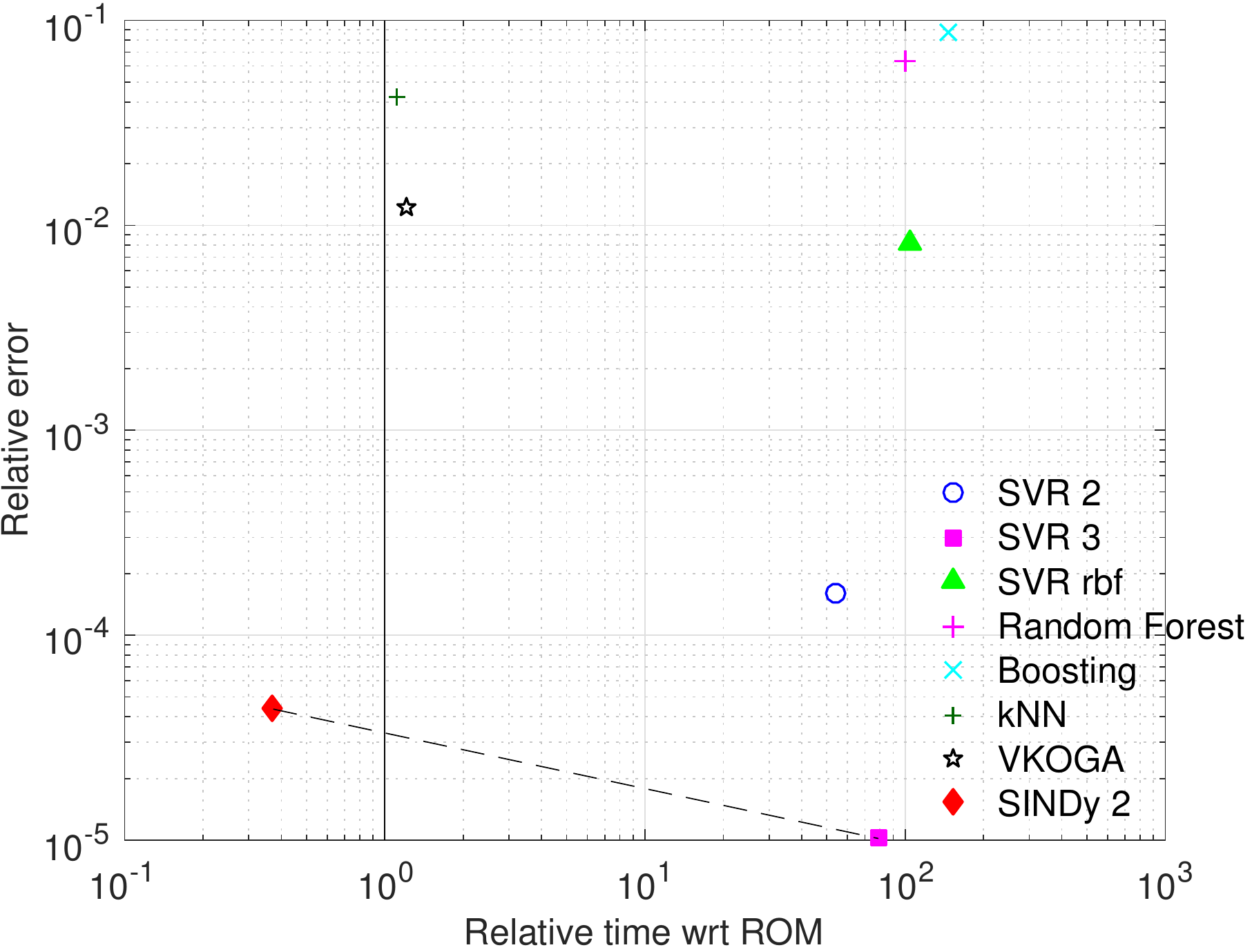}
		\put(-1,76){(b)}
		\put(-2,32){\colorbox{white}{\rotatebox{90}{\small{Relative error}}}}
		\put(32,-0.1){\colorbox{white}{\small{Relative time w.r.t. ROM}}}
	\end{overpic}
	\caption{Pareto frontier of relative error with respect to the relative running time using backward Euler for 1D inviscid Burgers' equation: (a) $\errorFOMave$ vs. $\tauFOM$ in FOM; (b) $\errorROMave$  vs. $\tauROM$ in ROM. }
	\label{BEpareto}
 \end{figure}
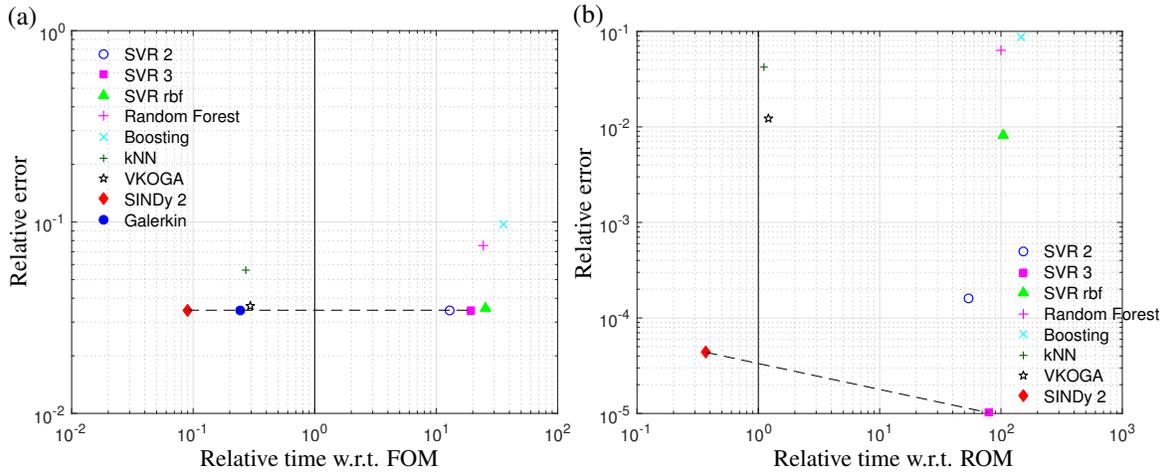

The following are the simulation results from backward Euler method with $\nTimesteps=800$. 
Figure~\ref{BEtime} plots the state-space error with respect to the FOM and ROM using the backward Euler integrator. As validation results predict, SVR3 and SINDy behave better than the other models, achieving a relative ROM error below $1e\text{-}4$ over the entire time domain, and the relative error in terms of FOM is well bounded.
Figure~\ref{BEpareto} plots the Pareto frontier error as a function of the relative running time using the backward Euler integrator. For differential models, the relative time is calculated using the less expensive approach, i.e. Newton's method. By comparison, SINDy requires much less relative time than SVR3, at a comparable level of relative error in both FOM and ROM. 
Table~\ref{tab:compMLmethods3} summarizes (i) online running time of all methods, (ii) mean time-integrated error with respect to the FOM, and (iii) mean time-integrated error with respect to ROMs using the backward Euler integrator.
For differentiable models, we show the computational time of both Newton's method and fixed point iteration in backward Euler. For those models that are non-differentiable, i.e. random forest, boosting and kNN, only the running time of the fixed-point iteration method is reported.
For a fair comparison, all the ROM and FOM solutions are computed at the verified backward Euler time step $\Delta{t} = 3.12e{\text -}2$. 
Note that the non-intrusive ROM, e.g. SINDy with Newton's method can accelerate the solver by $10.4 \times$ relative to the FOM and $2.5 \times$ compared to the Galerkin ROM at a relative error of $0.0346$ and $4.36e{\text -}5$ respectively.	
	
\begin{table}[ht]
\begin{center}
\caption{Comparison of different machine learning methods using backward Euler methods for 1D inviscid Burgers' equation: Newton's method (N); fixed-point iteration (FP). The running time of FOM and Galerkin is $3.886$s and $0.942$s respectively.} 
\label{tab:compMLmethods2}
\begin{tabular}{cccc}
\toprule
\textbf{Method}  & \textbf{Online running time (s)} & \textbf{R Err (w.r.t. FOM)} & \textbf{R Err (w.r.t. Galerkin)}  \\
\midrule
 SVR 2  FP & 93.365 &  0.0346 & 1.62e-4\\

 SVR 2  N  & 53.772  &  0.0346 & 1.62e-4 \\

SVR 3 FP  & 102.873 &  0.0346 & 1.02e-5  \\

SVR 3 N  & 78.935 & \textbf{0.0346} & \textbf{1.02e-5} \\

SVR rbf  FP  &  104.349  &  0.0353 & 8.12e-3  \\

SVR rbf  N &  104.548  &  0.0353& 8.12e-3 \\

Random Forest &  374.201 & 0.0756 & 0.0638 \\

Boosting     & 587.473 &  0.0980 & 0.0873 \\

kNN    &  1.108 &  0.0558 & 0.0427  \\

VKOGA FP  &  2.505  &  0.0365 & 0.0122  \\

VKOGA  N  &  1.217 & 0.0365 & 0.0122\\

SINDy FP & 0.534  & 0.0346 &  4.36e-5 \\

SINDy N   &  \textbf{0.373}   & \textbf{0.0346}  & 4.36e-5\\

Galerkin   &  0.942   &0.0346  &0 \\
\bottomrule
\end{tabular}
\end{center}
\end{table}

\begin{figure}
	\centering
	\begin{overpic}[width = 0.45\textwidth]{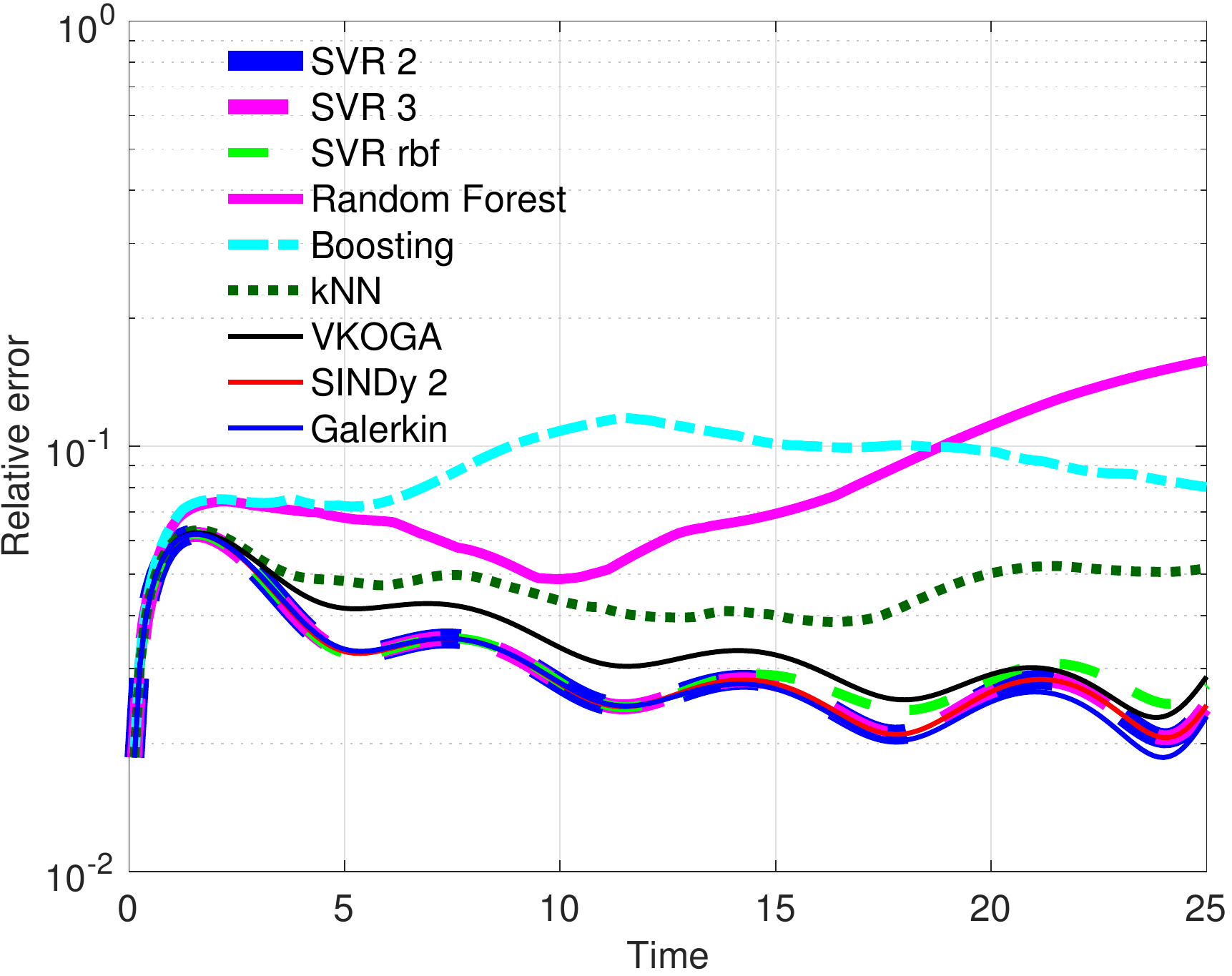}
		\put(-1,78){(a)}
		\put(49,-0.1){\colorbox{white}{\small{Time}}}
		\put(-2,32){\colorbox{white}{\rotatebox{90}{\small{Relative error}}}}
	\end{overpic}
	\begin{overpic}[width = 0.45\textwidth]{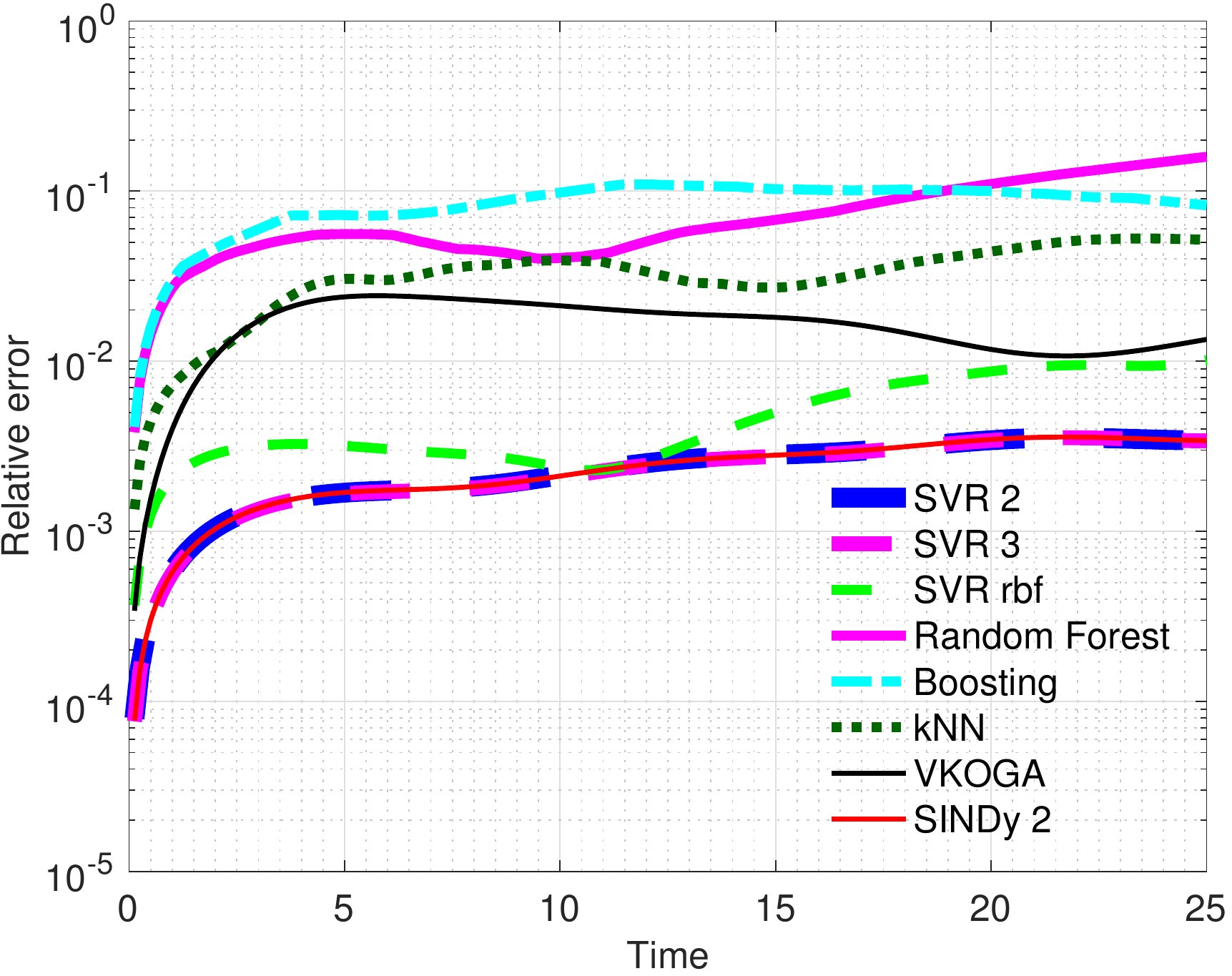}
		\put(-1,78){(b)}
		\put(49,-0.1){\colorbox{white}{\small{Time}}}
		\put(-2,32){\colorbox{white}{\rotatebox{90}{\small{Relative error}}}}
	\end{overpic}
	\caption{Runge-Kutta for 1D inviscid Burgers' equation: time evolution of relative error: (a) $\errorFOM$ in FOM; and (b) $\errorROM$ in ROM.}
	\label{RKtime}
 \end{figure}

\begin{figure}
	\centering
	\vspace{.3in}
	\begin{overpic}[width = 0.45\textwidth]{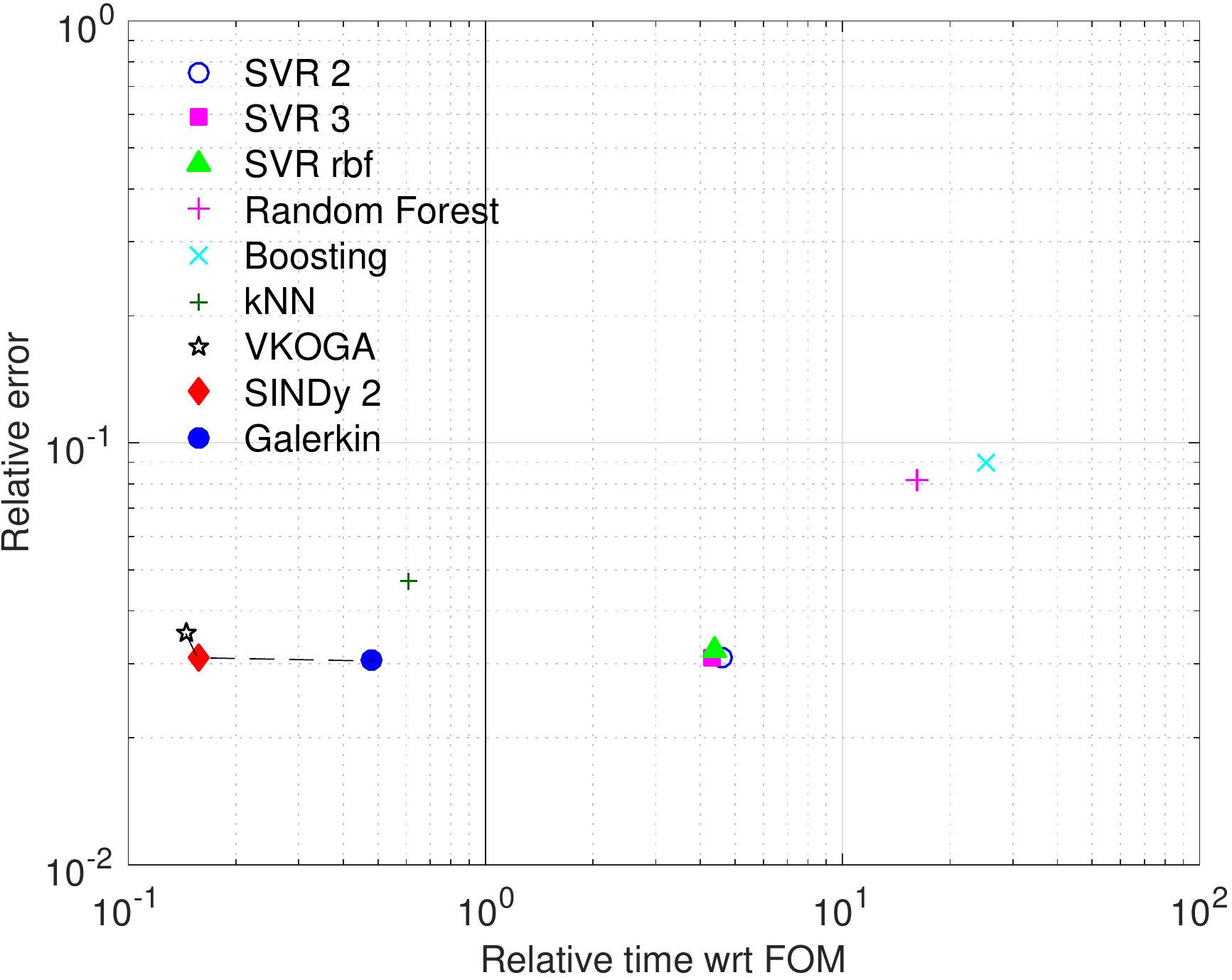}
		\put(-1,79){(a)}
		\put(-2,32){\colorbox{white}{\rotatebox{90}{\small{Relative error}}}}
		\put(32,-0.1){\colorbox{white}{\small{Relative time w.r.t. FOM}}}
	\end{overpic}
	\begin{overpic}[width = 0.45\textwidth]{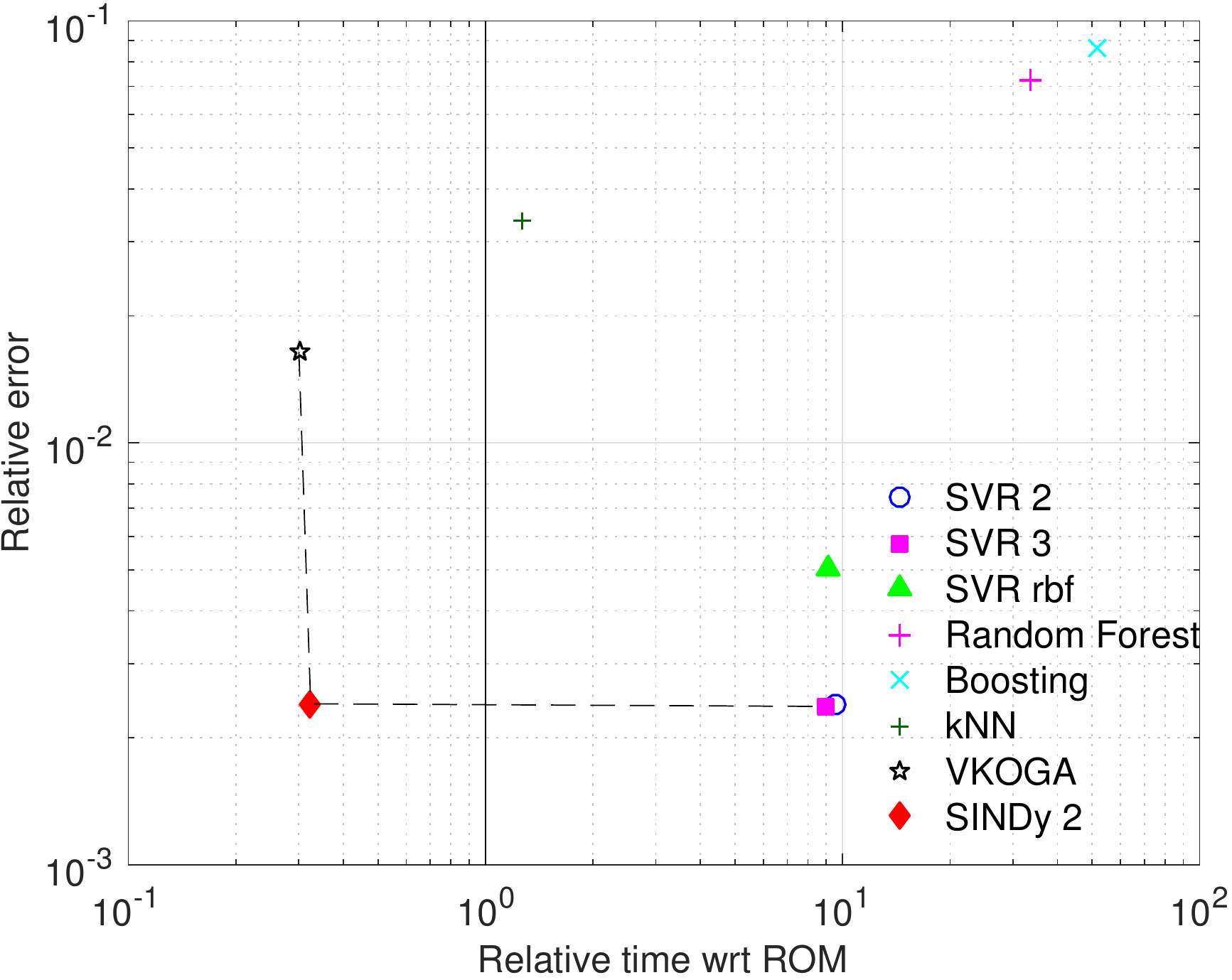}
		\put(-1,79){(b)}
		\put(-2,32){\colorbox{white}{\rotatebox{90}{\small{Relative error}}}}
		\put(32,-0.1){\colorbox{white}{\small{Relative time w.r.t. ROM}}}
	\end{overpic}
	\caption{
	Pareto frontier of relative error with respect to the relative running time using 4th-order Runge-Kutta for 1D inviscid Burgers' equation: (a) $\errorFOMave$ vs. $\tauFOM$ in FOM; (b) $\errorROMave$  vs. $\tauROM$ in ROM.}
	\label{RKpareto}
 \end{figure}
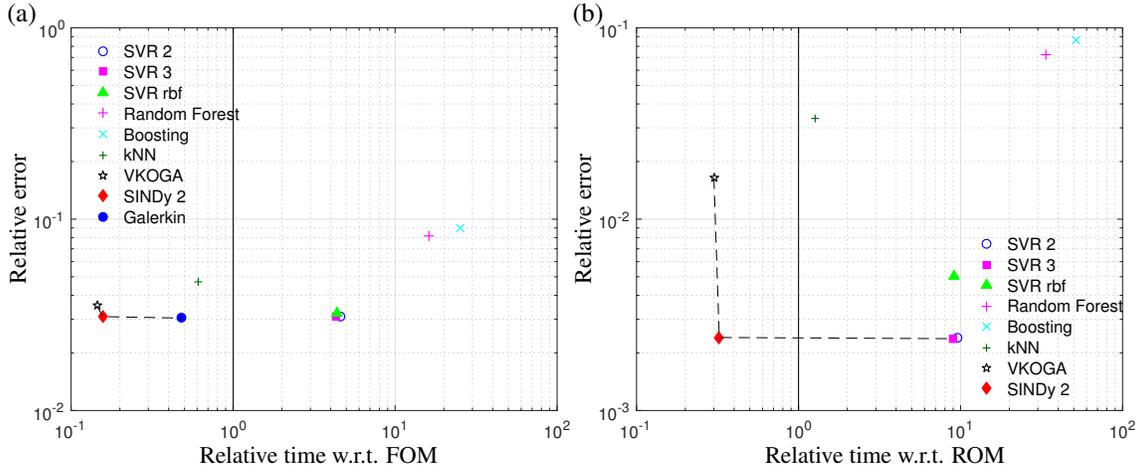

We examine simulation results from 4th-order Runge-Kutta method with  $\nTimesteps=200$. 
Figure~\ref{RKtime} shows the state-space error with respect to the FOM and ROM using the Runge-Kutta integrator. SVR2, SVR3 and SINDy have a comparable performance, and result in a bigger ROM error relative to the backward Euler solver. We notice that the random forest model begins to diverge after $\timevar>10$ in the explicit scheme. This can be explained by the corresponding performance in model evaluation in Appendix~\ref{app1}. 
Figure~\ref{RKpareto} plots the Pareto frontier error with respect to the relative running time using the backward Euler integrator. VKOGA has the smallest relative time in both the FOM and ROM comparison. SINDy requires slightly more running time while the accuracy outperforms VKOGA. 
Table~\ref{tab:compMLmethods3} summarizes the online running time of all methods, the mean time-integrated error with respect to the FOM, and the mean time-integrated error with respect to the Galerkin ROM using the Runge-Kutta solver.
For a fair comparison, all the ROM and FOM solutions are computed at the verified Runge-kutta time step. As shown in the table, SVR based models, e.g. SVR2 and SVR3, yield the smallest relative errors, however the computational cost is more expensive than the FOM.
Note that the non-intrusive ROM (VKOGA) can speed up the solver by $6.9 \times$ relative to the FOM and $3.3 \times$ over the Galerkin ROM at a relative error of $0.0353$ and $0.0164$ respectively.	
	 
\begin{table}[ht]
\begin{center}
\caption{Comparison of different machine learning methods using $4$th-order Runge-Kutta for 1D inviscid Burgers' equation. The running time of FOM and Galerkin is $0.401$s and  $0.194$s respectively.} 
\label{tab:compMLmethods3}
\begin{tabular}{cccc}
\toprule
\textbf{Method}  & \textbf{Online running time (s)} & \textbf{R Err (w.r.t. FOM)} & \textbf{R Err (w.r.t. Galerkin)}  \\
\midrule
 SVR 2  &   1.854 &  0.0309 & 2.42e-3  \\

SVR 3 & 1.744  & \textbf{0.0310} & \textbf{2.42e-3}  \\

SVR rbf    &  1.773 &  0.0321  & 5.03e-3  \\

Random Forest   & 6.513  &0.0817  & 0.0726 \\

Boosting     &  10.076  &  0.0904 & 0.0858 \\

kNN    &  0.244 &  0.0467 & 0.0336\\

VKOGA &  \textbf{0.058}  &  0.0353  &0.0164 \\

SINDy  &  0.063 & \textbf{0.0310} &  \textbf{2.42e-3} \\

Galerkin  &  0.194  & 0.0304 &  0 \\
\bottomrule
\end{tabular}
\end{center}
\end{table}

\subsection{2D convection--diffusion}\label{sec4.2}
\begin{figure}
	\centering
	\vspace{.3in}
	\begin{overpic}[width = 0.45\textwidth]{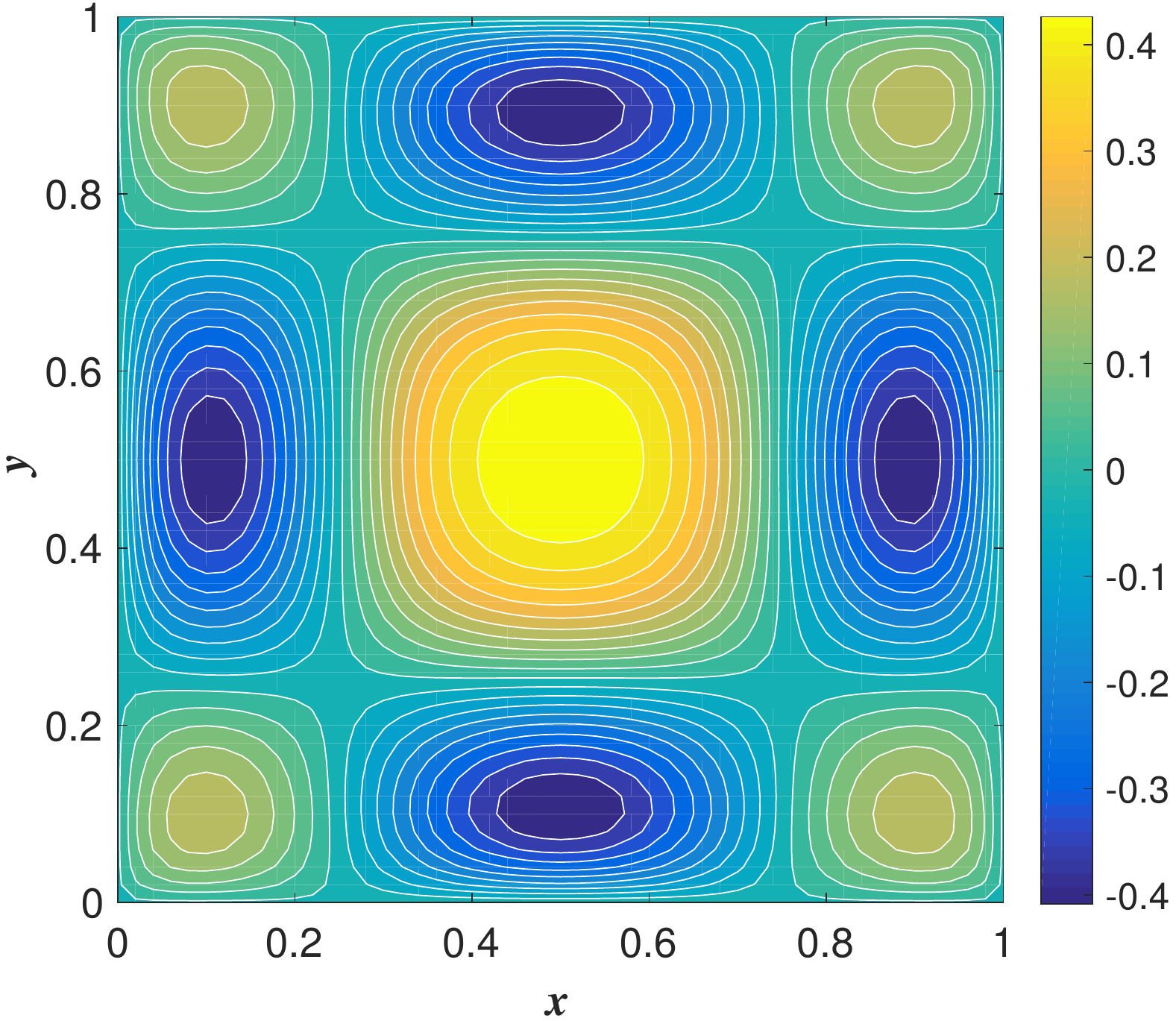}
		\put(-1,79){(a)}
	\end{overpic}
	\begin{overpic}[width = 0.45\textwidth]{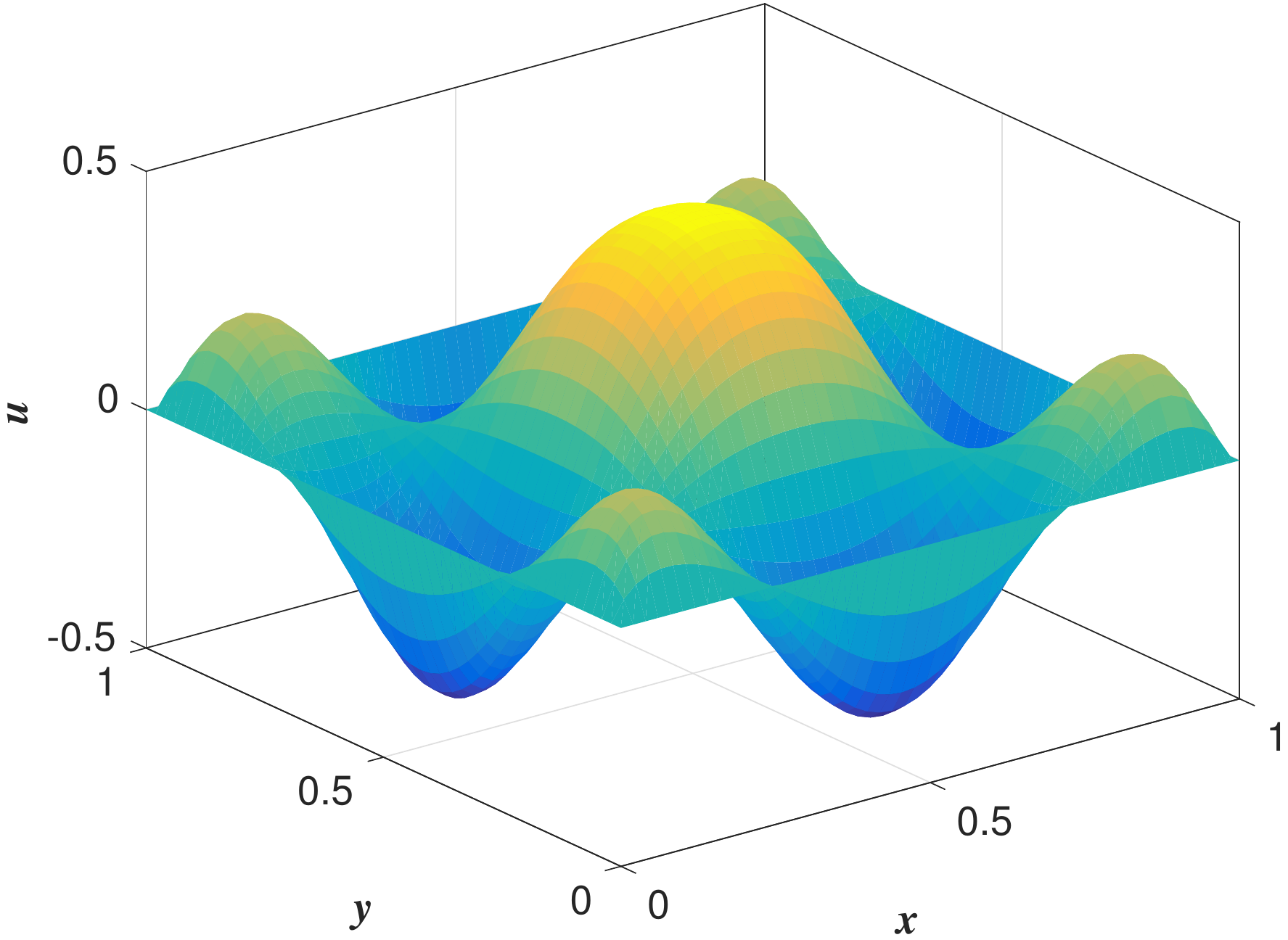}
		\put(-1,79){(b)}
	\end{overpic}
	\caption{Solution profile of 2D convection--diffusion equation, $u(x, y, t= 2)$ with input parameter $\mu_1 = \mu_2 =9.5$.}
	  \label{fig:sol_parabolic2}
 \end{figure}
 
 \begin{figure}
	\centering
	\vspace{.3in}
	\begin{overpic}[width = 0.95\textwidth]{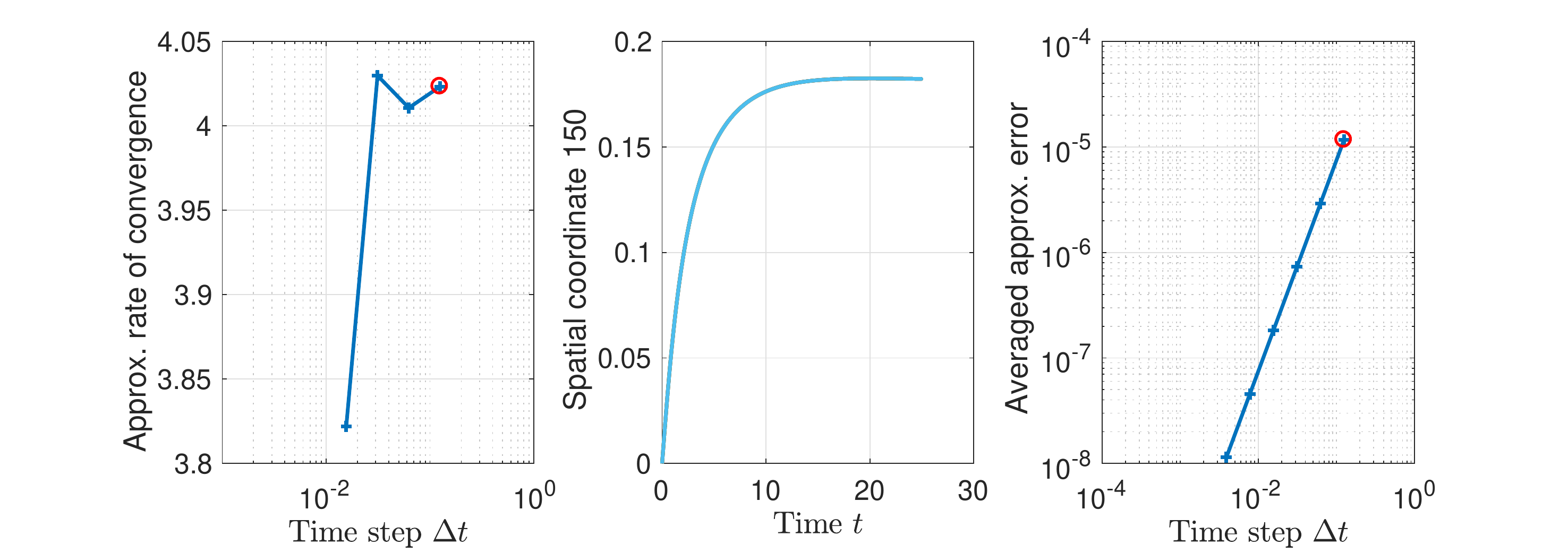}
		\put(3,32){(a)}
	\end{overpic}\\
	\begin{overpic}[width = 0.95\textwidth]{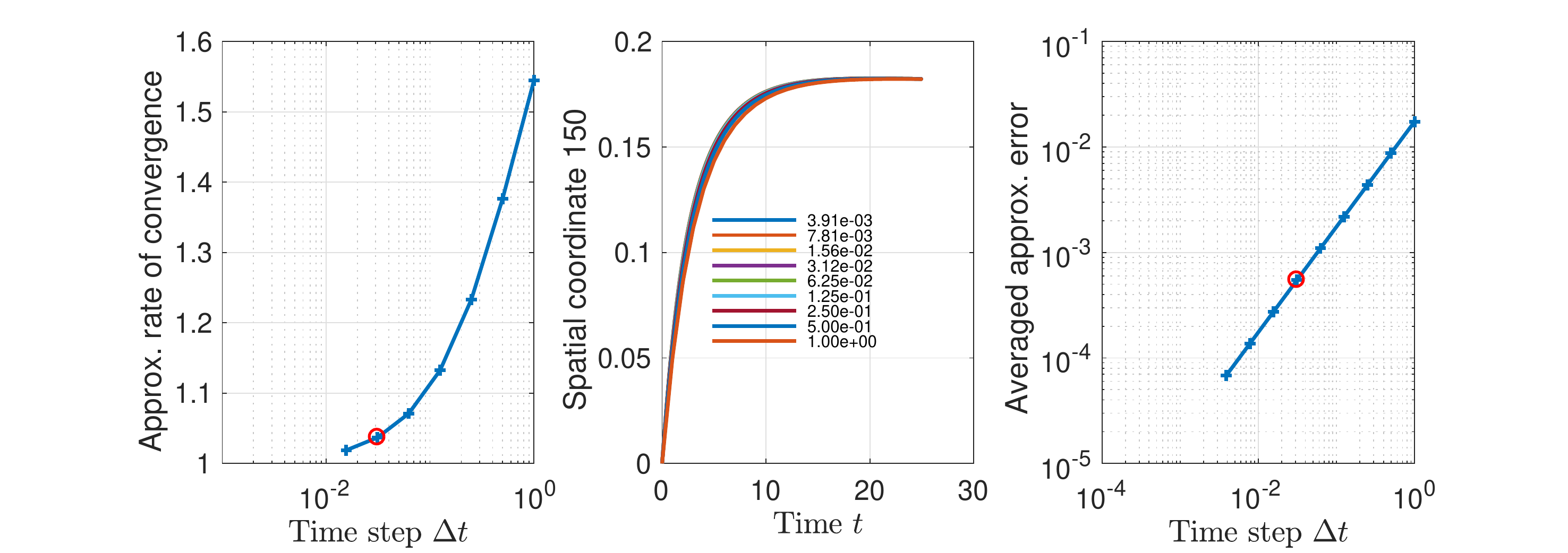}
		\put(3,32){(b)}
	\end{overpic}
	\caption{Timestep verification for 2D convection--diffusion equation: (a) Runge-Kutta: we select a time step of $1.25e{\text -}1$, as it leads to an approximated convergence rate of $4$ and an averaged approximated error of $2e{\text -}5$ for the selected state. (b) backward Euler integrator: we select a time step of $3.12e{\text -}2$, as it leads to an approximated convergence rate of 1 and an approximated error of $5e{\text -}4$ for the selected state.}
	\label{FigTimestep2D}
\end{figure}
We consider a 2D parametric nonlinear hear equation. Given a state variable $u = u(x, y, t)$, the governing equation is described as
\begin{subequations}\label{eq:2D heat}
\begin{align}
    \frac{\partial u(x, y, t)}{\partial t} &= -\mu_0 \bigtriangledown^2  u - \frac{\mu_0 \mu_1}{\mu_2} \left( e^{\mu_2 u} - 1\right)  + \cos(2\pi x) \cos (2\pi y), \tag{\ref{eq:2D heat}}\\
    u(x, y, 0) &= 0.
\end{align}
\end{subequations}
The parameters are given by 
$\mu_0 = 0.01$, and $(\mu_1, \mu_2) \in [9, 10]^2$. The spatial domain is $[0,1]^2$ and Dirichlet boundary conditions are applied. The FOM uses a finite difference discretization with $51 \times 51$ grid points. The full time domain is $[0, 2]$ and we evaluate both the backward Euler and Runge-Kutta methods for time integration with uniform time steps. Figure~\ref{fig:sol_parabolic2} shows the solution profile at $t = 2$ with input parameter $(\mu_1, \mu_2) = (9.5, 9.5)$.
\subsubsection{Data collection}
Similar to the 1D case, first we investigate the appropriate time step $\Delta t$ for solving the ODE. We collect the solutions of a sequence number of time steps $\nTimesteps = [25, 50, 100, 200, 400,  800, 1600, 3200, 6400]$ for (i) explicit Runge-Kutta and (ii) implicit backward-Euler integrator. The verification results in Figure~\ref{FigTimestep2D} shows that $\nTimesteps = 200$ is a reasonable number of time steps to use for Runge-Kutta and $\nTimesteps=800$ for backward Euler method.
During the offline stage, we run four full simulations corresponding to the corner parameters of the space $[9, 10]^2$. Then, the training data are sampled from a Latin-hypercube for better covering the parameter space. In the sampling, $\ntrain$ and  $\nvalidation$ instances of the state, time and parameters are generated following the criterion that the minimum distances between the data points are maximized.  We use the default size of the training set $\ntrain = 1000$ and of the validation set $\nvalidation = 500$. Then the reduced vector field $\velocityRed$ is computed for each input pairs $(\mystateRed,\timevar;\params)$. In the training and validation stage, we regress the reduced vector field $\velocityRed$ by the input $(\mystateRed,\timevar;\params)$; in the test stage, we evaluate the ROM. The parameters are fixed to be $(\mu_1, \mu_2) = (9.5, 9.5)$.
\subsubsection{Model validation}
We report the performance of SVR(2$^{nd}$, 3$^{rd}$ poly and rbf), kNN, Random Forest, Boosting, VKOGA, and SINDy as regression models to approximate reduced velocity. 
  In particular, as in Section~\ref{Section4.1.3}, for each regression method, we change the model hyperparameters and plot the relative training and validation error. The relative error is defined by Equation~(\ref{err}).
 Similarly, we plot the learning curve of each regression method and compare the performance of each model on training and validation data over a varying number of training instances in Appendix~\ref{app2}. We aim to balance bias and variance in each regression model, by properly choosing hyperparameters and the number of training instances.

\subsubsection{Simulation of the surrogate ROM}
\begin{figure}
	\centering
	\begin{overpic}[width = 0.45\textwidth]{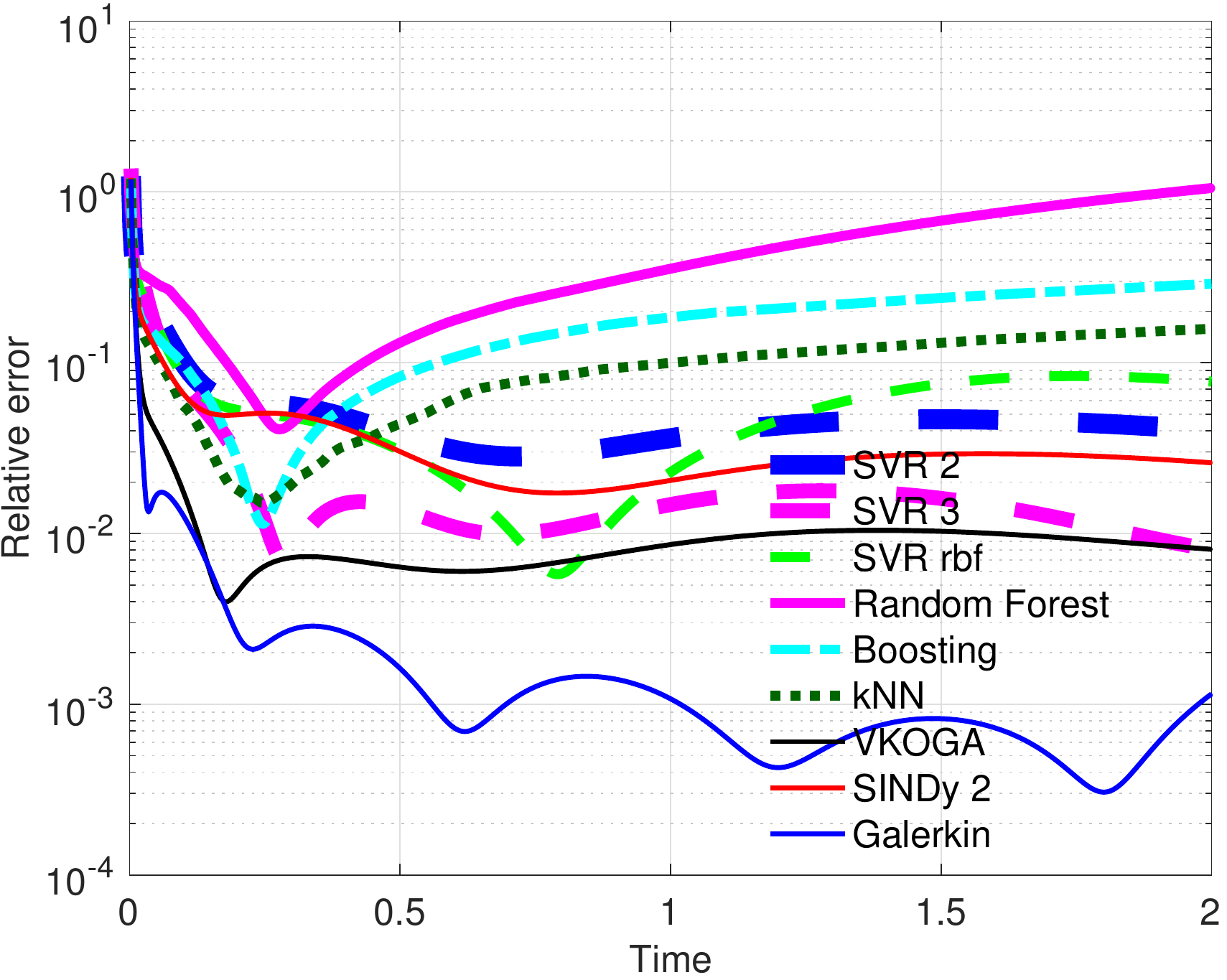}
		\put(-1,78){(a)}
		\put(49,-0.1){\colorbox{white}{\small{Time}}}
		\put(-2,32){\colorbox{white}{\rotatebox{90}{\small{Relative error}}}}
	\end{overpic}
	\begin{overpic}[width = 0.45\textwidth]{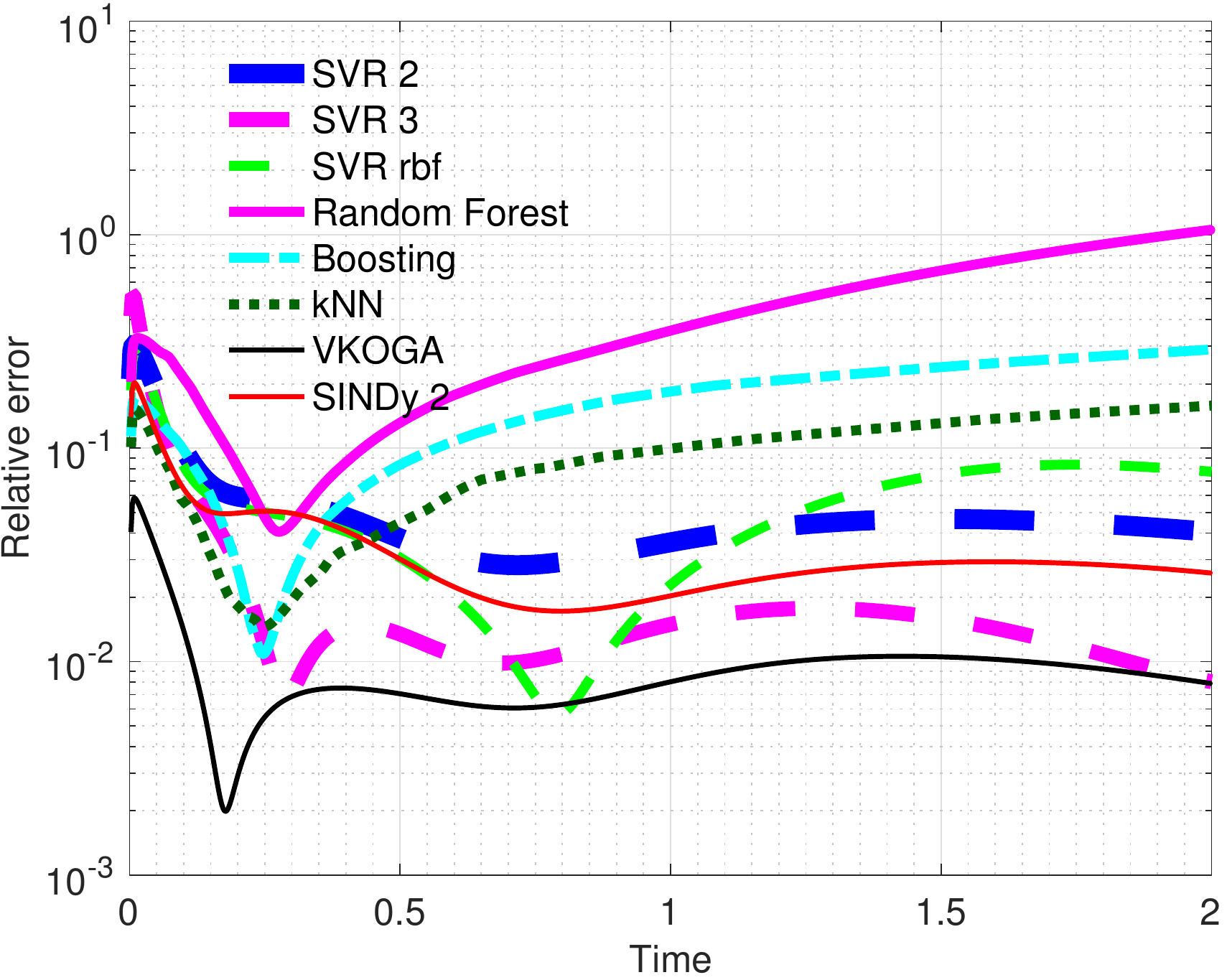}
		\put(-1,78){(b)}
		\put(49,-0.1){\colorbox{white}{\small{Time}}}
		\put(-2,32){\colorbox{white}{\rotatebox{90}{\small{Relative error}}}}
	\end{overpic}
	\caption{Backward Euler for 2D convection--diffusion equation: time evolution of relative error: (a) $\errorFOM$ in FOM; and  (b) $\errorROM$ in ROM.}
	\label{BEtime2}
 \end{figure}
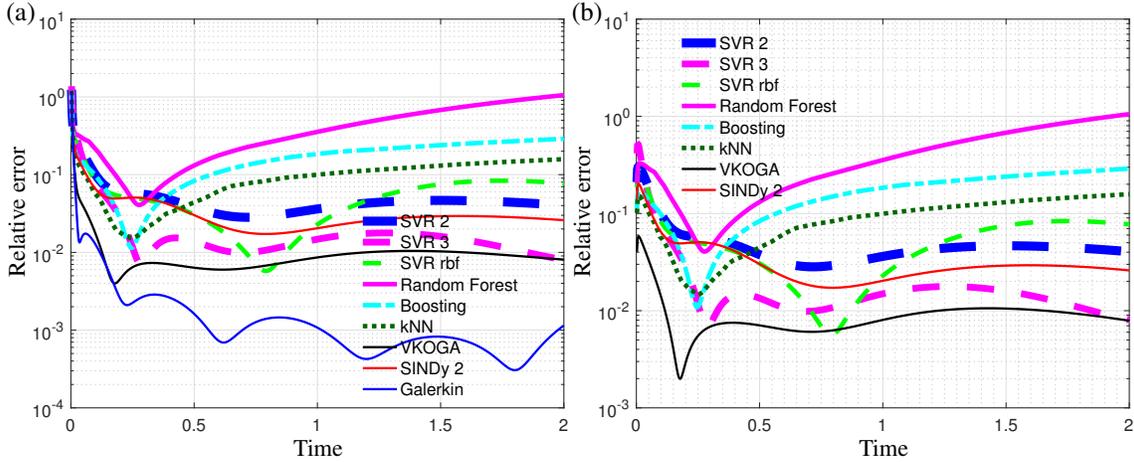

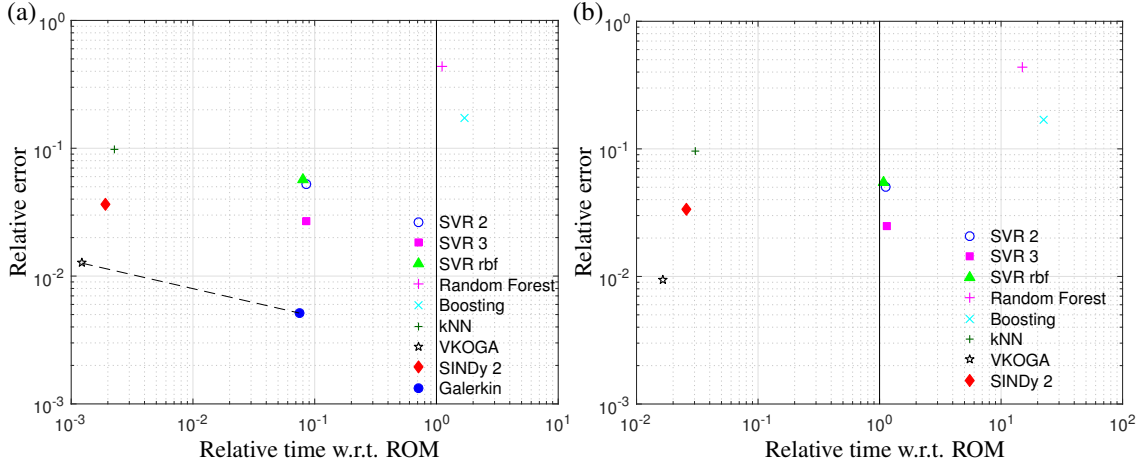
\begin{figure}
	\centering
	\begin{overpic}[width = 0.45\textwidth]{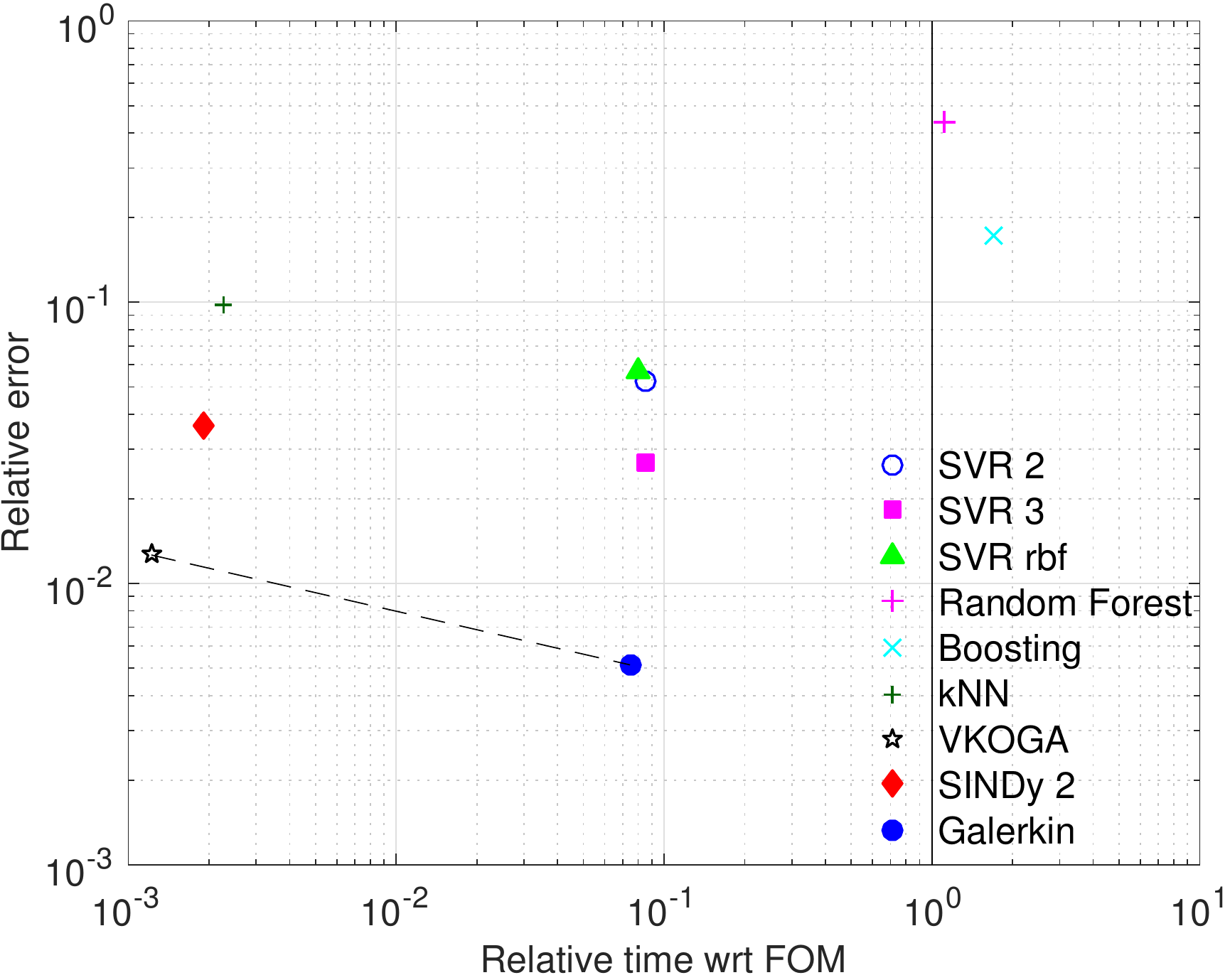}
		\put(-1,78){(a)}
		\put(-2,32){\colorbox{white}{\rotatebox{90}{\small{Relative error}}}}
		\put(32,-0.1){\colorbox{white}{\small{Relative time w.r.t. ROM}}}
	\end{overpic}
	\begin{overpic}[width = 0.45\textwidth]{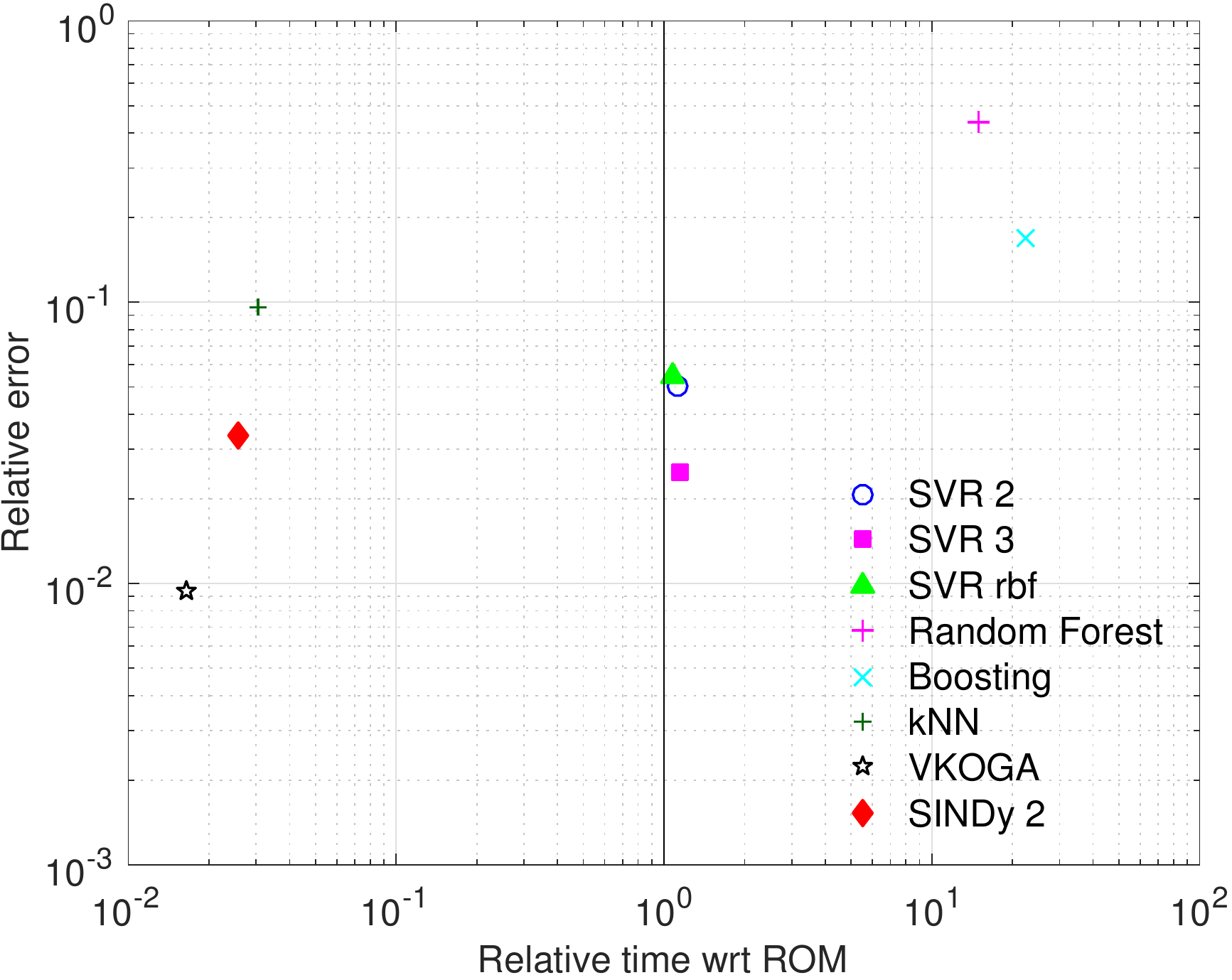}
		\put(-1,78){(b)}
		\put(-2,32){\colorbox{white}{\rotatebox{90}{\small{Relative error}}}}
		\put(32,-0.1){\colorbox{white}{\small{Relative time w.r.t. ROM}}}
	\end{overpic}
	\caption{Pareto frontier of relative error with respect to the relative running time using backward Euler for 2D convection--diffusion equation: (a) $\errorFOMave$ vs. $\tauFOM$ in FOM; (b) $\errorROMave$  vs. $\tauROM$ in ROM. }
	\label{BEpareto2}
 \end{figure}

We can now solve this 2D problem using the surrogate model along the trajectory of the dynamical system. After applying time integration to the regression-based ROM, we compute the relative error of the proposed models as a function of time. As in Section~\ref{Section4.1.4}, we investigate both the backward Euler and Runge-Kutta integrators.
The following are the simulation results from backward Euler method with  $\nTimesteps=800$. 
Figure~\ref{BEtime2} plots the state-space error with respect to FOM and ROM using the backward Euler integrator. VKOGA outperforms the other models, achieving a relative ROM error below $6e\text{-}2$ over the entire time domain, and the accuracy is closest to Galerkin ROM. 
Figure~\ref{BEpareto2} plots the Pareto frontier error as a function of the relative running time using the backward Euler integrator. VKOGA performs best in terms of both accuracy and time efficiency, when comparing with the Galerkin ROM.
Table~\ref{tab:compMLmethods6} represents online running time of all methods, the mean time-integrated error with respect to FOM, and the mean time-integrated error with respect to the Galerkin ROM using the backward Euler integrator for the 2D convection--diffusion equation.
For a fair comparison, all the ROM and FOM solutions are computed at the verified backward Euler time step. 
Note that the non-intrusive ROM, e.g. VKOGA with Newton's method, can improve the solve time by three orders of magnitude over the FOM and $111.2 \times$ compared to the Galerkin ROM at a relative error of $0.0059$ and $0.0041$ respectively.		
\begin{table}
\begin{center}
\caption{Comparison of different machine learning methods using the Backward Euler integrator for 2D convection--diffusion equation: Newton's method (N); FP-fixed-point iteration (FP). The running time of FOM and Galerkin is $209.393$s and $16.837$s respectively.} 
\label{tab:compMLmethods6}
\begin{tabular}{cccc}
\toprule
\textbf{Method}  & \textbf{Online running time (s)} & \textbf{R Err (w.r.t. FOM)} & \textbf{R Err (w.r.t. Galerkin)}  \\
\midrule
 SVR 2  FP &  13.510 &  0.0074 & 0.0061\\

 SVR 2  N  & 10.510  &   0.0074 &  0.0061\ \\

SVR 3 FP  & 13.873 &  0.0045 & 0.0028  \\

SVR 3 N  & 10.135 & 0.0045 & 0.0028 \\

SVR rbf  FP  &  13.349  &  0.0181& 0.0169  \\

SVR rbf  N &  10.405  &  0.0181 & 0.0169\\

Random Forest & 120.674 & 0.0332 & 0.0326  \\

Boosting     & 190.632  &  0.0147 &  0.0135  \\

kNN    & 0.435 &  0.0149 & 0.0142   \\

VKOGA FP  &  2.505  & 0.0059& 0.0041  \\

VKOGA  N  & \textbf{0.151} & \textbf{0.0059} & \textbf{0.0041}\\

SINDy FP & 0.534  &  0.0066 &  0.0054\\

SINDy N   & 0.236   &  0.0066 & 0.0054\\

Galerkin   &  0.942   & 0.0029  &0 \\
\bottomrule
\end{tabular}
\end{center}
\end{table}
  
\begin{figure}
	\centering
	\vspace{.3in}
	\begin{overpic}[width = 0.45\textwidth]{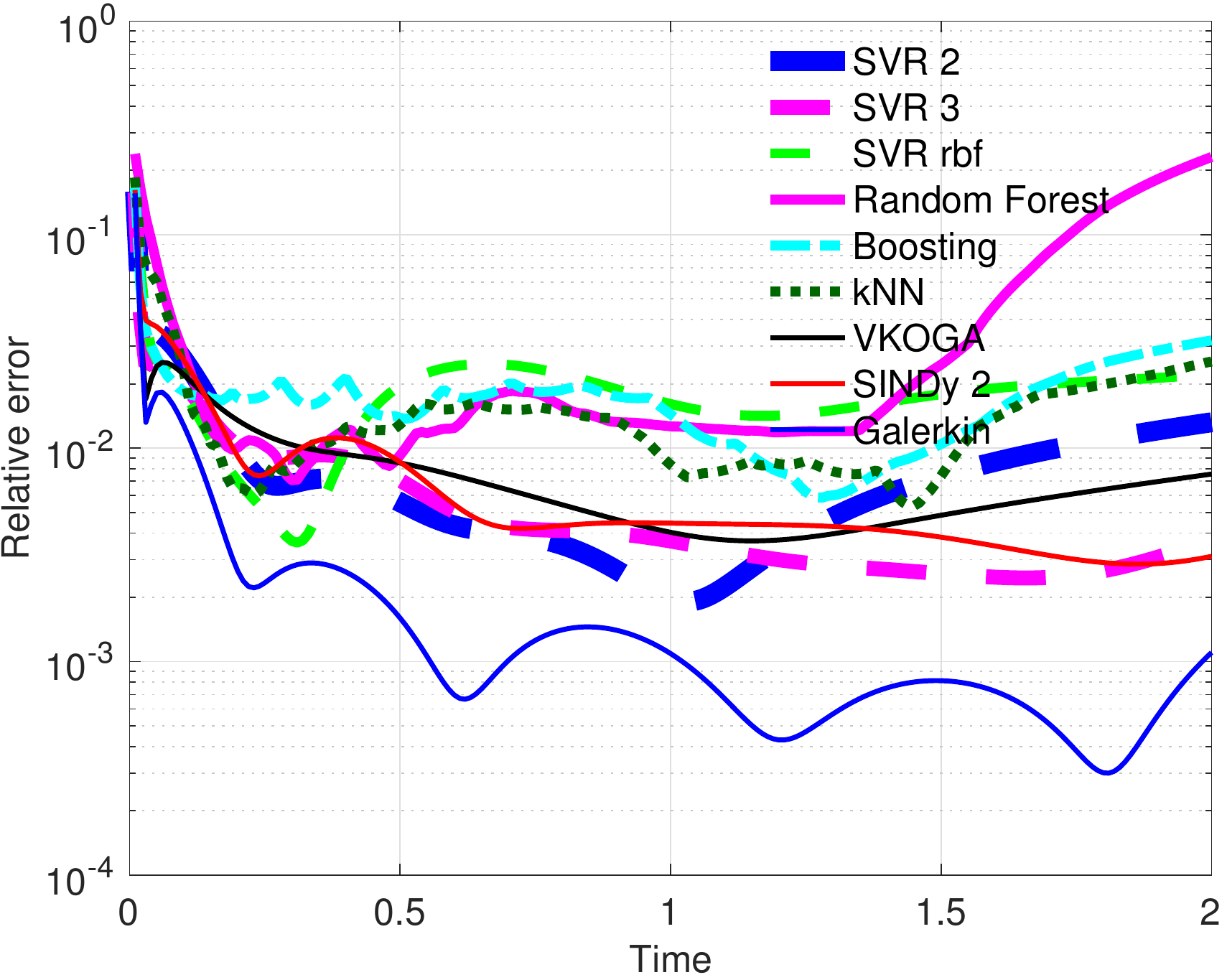}
		\put(-1,78){(a)}
		\put(-2,32){\colorbox{white}{\rotatebox{90}{\small{Relative error}}}}
		\put(49,-0.1){\colorbox{white}{\small{Time}}}
	\end{overpic}
	\begin{overpic}[width = 0.45\textwidth]{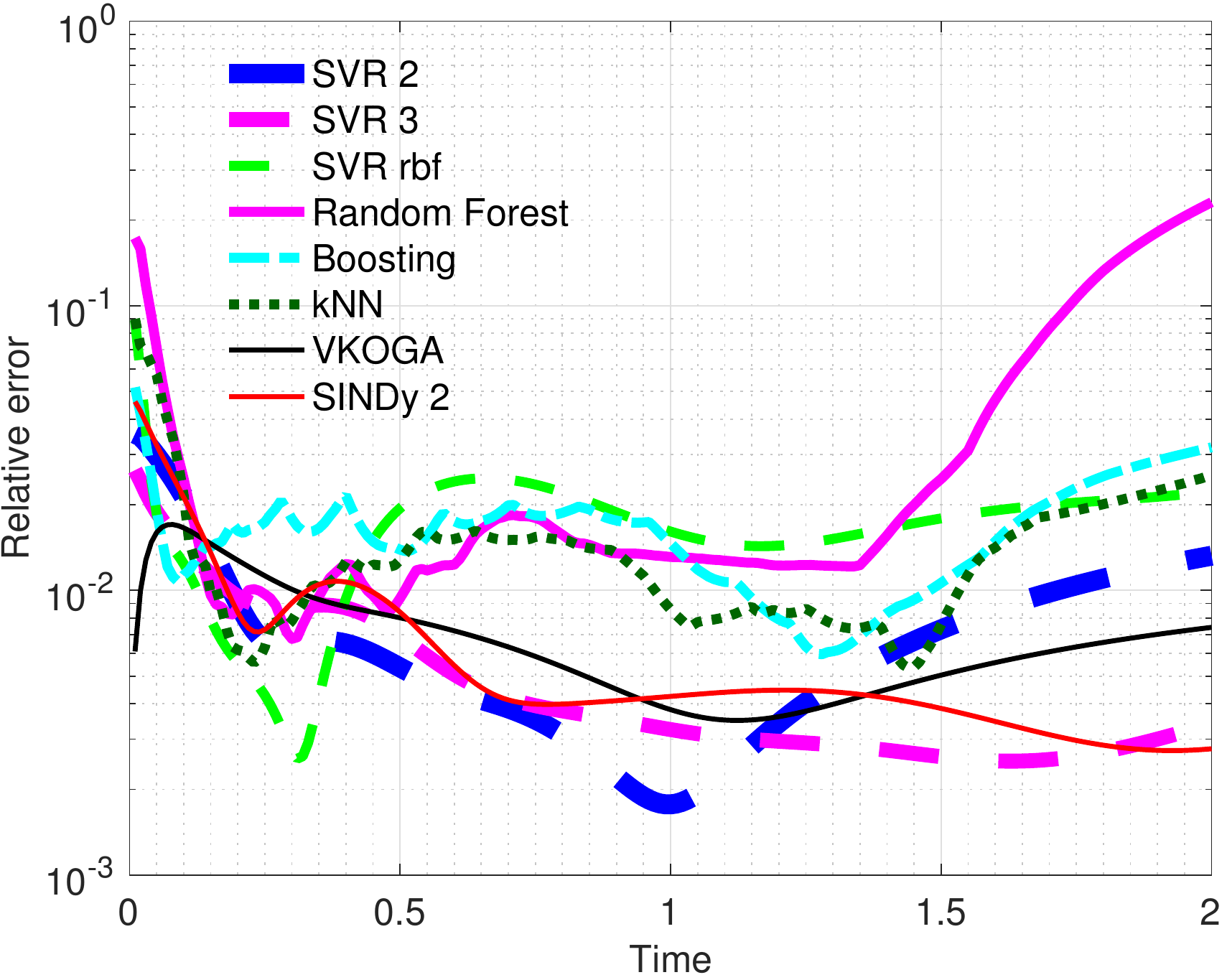}
		\put(-1,78){(b)}
		\put(-2,32){\colorbox{white}{\rotatebox{90}{\small{Relative error}}}}
		\put(49,-0.1){\colorbox{white}{\small{Time}}}
	\end{overpic}
	\caption{Runge-Kutta for 2D convection--diffusion equation: time evolution of relative error: (a) $\errorFOM$ in FOM; and  (b) $\errorROM$ in ROM.}
	\label{RKtime2}
 \end{figure}
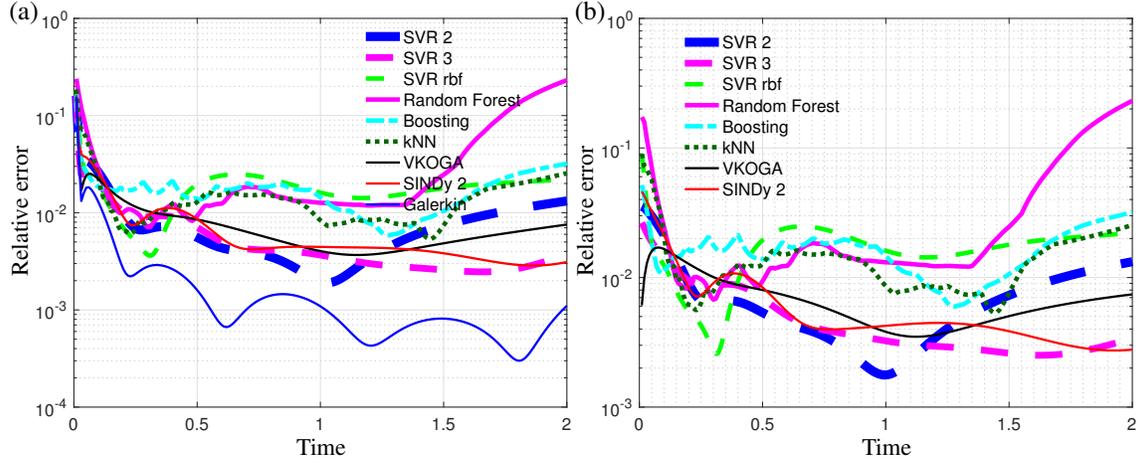

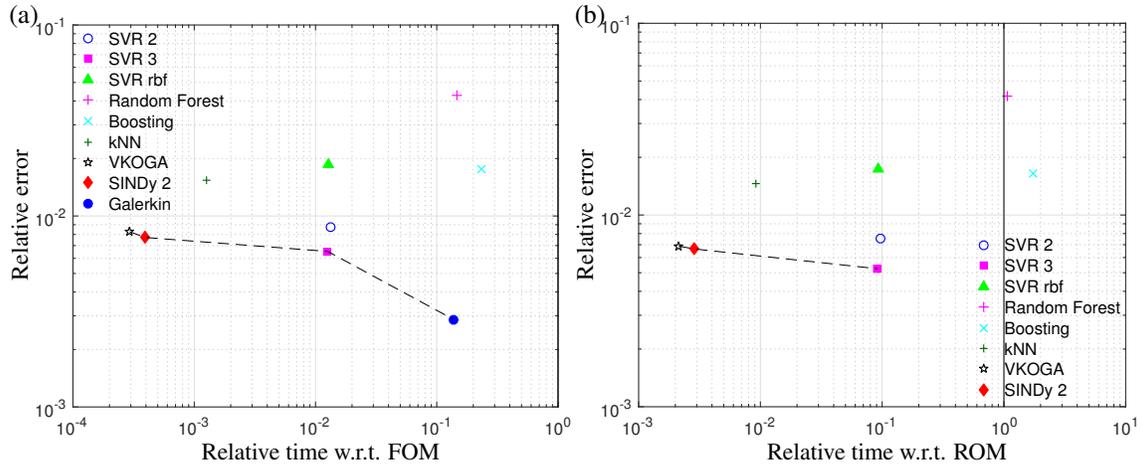
\begin{figure}
	\centering
	\begin{overpic}[width = 0.45\textwidth]{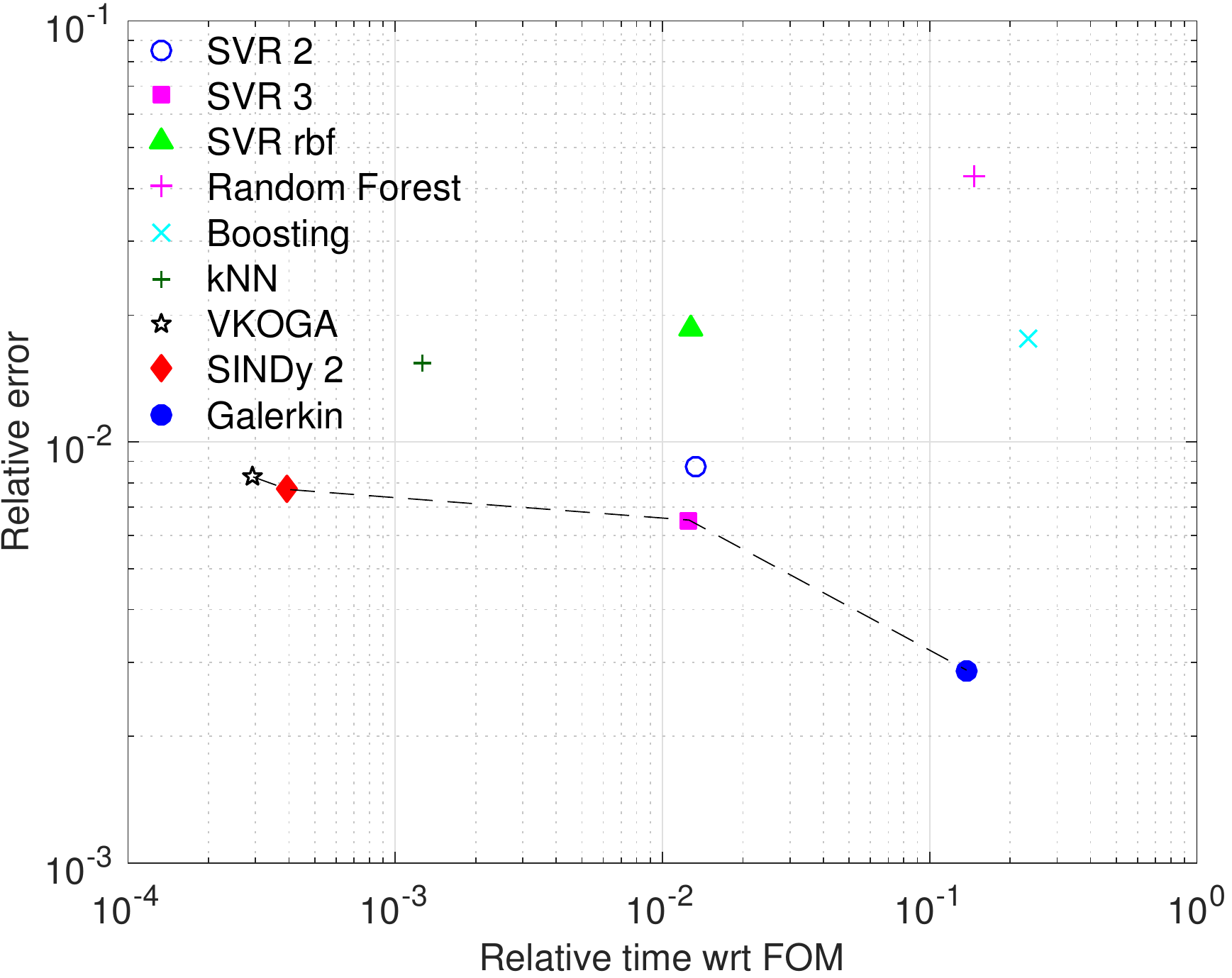}
		\put(-1,78){(a)}
		\put(-2,32){\colorbox{white}{\rotatebox{90}{\small{Relative error}}}}
		\put(32,-0.1){\colorbox{white}{\small{Relative time w.r.t. FOM}}}
	\end{overpic}
	\begin{overpic}[width = 0.45\textwidth]{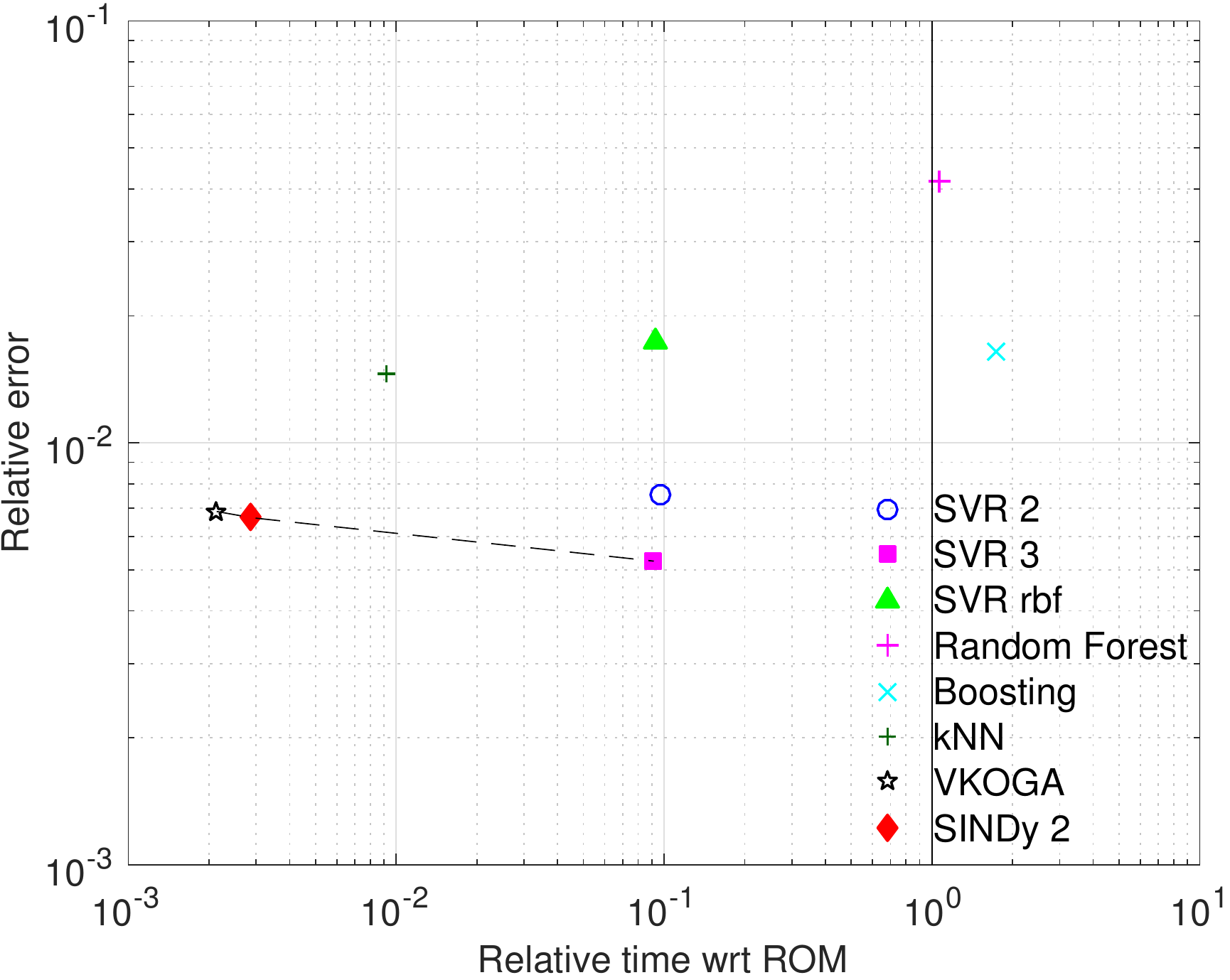}
		\put(-1,78){(b)}
		\put(-2,32){\colorbox{white}{\rotatebox{90}{\small{Relative error}}}}
		\put(32,-0.1){\colorbox{white}{\small{Relative time w.r.t. ROM}}}
	\end{overpic}
	\caption{Pareto frontier of relative error with respect to the relative running time using 4th-order Runge-Kutta for 2D convection--diffusion equation: (a) $\errorFOMave$ vs. $\tauFOM$ in FOM; (b) $\errorROMave$  vs. $\tauROM$ in ROM.}
	\label{RKpareto2}
 \end{figure}

We examine the simulation results from the 4th-order Runge-Kutta method with  $\nTimesteps=200$. 
Figure~\ref{RKtime2} shows the state-space error with respect to FOM and ROM using the Runge-Kutta integrator. SVR2, SVR3, VKOGA, and SINDy have a comparable performance, and result in a smaller ROM error relative to in the backward Euler solver. We notice that the random forest, boosting models and kNN begin to diverge quickly in the second half of the time domain. 
Figure~\ref{RKpareto2} plots the Pareto frontier error as a function of the relative running time using the backward Euler integrator. VKOGA and SINDy outperform the other models in terms of both computation accuracy and time cost. The relative error compared to Galerkin ROM and FOM is below $1e{\text -}2$.
Table~\ref{tab:compMLmethods7} presents online running time of all methods, the mean time-integrated error with respect to the FOM, and the mean time-integrated error with respect to the Galerkin ROM using the Runge-Kutta integrator for the 2D convection--diffusion equation.
For a fair comparison, all the ROM and FOM solutions are computed at the verified backward Euler time step. 
Note that the computational efficiency of the non-intrusive ROM performs significantly better than Galerkin ROM. VKOGA using Runge-Kutta can speed up the solver $3406.9 \times$ that of the FOM and $468.3 \times$ that of the Galerkin ROM at a relative error of $0.0083$ and $0.0069$.  SINDy using Runge-Kutta can accelerate the solver $2524.1 \times$ over the FOM and $347.0 \times$ compared to the Galerkin ROM at a relative error of $0.0077$ and $0.0066$ respectively.	 	
	
\begin{table}
\begin{center}
\caption{Comparison of different machine learning methods using $4$th-order Runge-Kutta for 2D convection--diffusion equation. The running time of FOM and Galerkin is $138.319$s and $19.012$s respectively.} 
\label{tab:compMLmethods7}
\begin{tabular}{cccc}
\toprule
\textbf{Method}  & \textbf{Online running time (s)} & \textbf{R Err (w.r.t. FOM)} & \textbf{R Err (w.r.t. Galerkin)}  \\
\midrule
 SVR 2  &   1.850 &  0.0087 & 0.0076  \\

SVR 3 & 1.740  & 0.0065 & 0.0052  \\

SVR rbf    & 1.777   &  0.0185  & 0.0174  \\

Random Forest   & 20.313  &0.0426  & 0.0419 \\

Boosting     & 32.668 &  0.0177 & 0.0166 \\

kNN    &  0.175 &  0.0153 & 0.0146 \\

VKOGA &  \textbf{0.041}  &  0.0083  &0.0069 \\

SINDy  &  0.055 & \textbf{0.0077} &  \textbf{0.0066} \\

Galerkin  &  19.012  & 0.0029 &  0 \\
\bottomrule
\end{tabular}
\end{center}
\end{table}
 
\section{Conclusions}\label{sec5}
In this study, we demonstrate the effectiveness of a data-driven model reduction method for solving two types of parametric PDEs non-intrusively, in both explicit and implicit time integration schemes. The approach successfully avoids the cost and intrusiveness of querying the full-order model and reduced-order model in the simulation, by approximating low-dimensional operators using regression methods. In the offline stage, we train the regression models using state-of-the art techniques from machine learning and specific dynamical learning methods in a reduced order architecture; in the online stage, the surrogate ROM then solves problems more efficiently (orders of magnitude in speedup) using the learned reduced operators. The proposed approach speeds up the full-order model simulation by one order of magnitude in the 1D Burgers' equation and three order of magnitude in the 2D convection-diffusion equation.  Among the machine learning models we evaluate, VKOGA and SINDy distinguish themselves from the other models, delivering superior cost vs. error trade-off. 

Further work involves a number of important extensions and directions that arise out of this work. First, it will be interesting to investigate the effectiveness on solving a different types of PDE or a more complex nonlinear dynamical systems, e.g. Euler’s equation. Morerover, for nonlinear Lagrangian dynamical systems, we need to develop structure-preserving approximations for all reduced Lagrangian ingredients in the model reduction. Rather than apply a Galerkin projection to obtain the ROM, one can alternatively employ a least-square Petrov-Galerkin (LSPG)~\citep{CarlbergGappy, Carlberg2013jcp, Carlberg2017jcp}, which requires a regression method predicting non-negative values. 
The growing intersection of dynamical systems and data science are driving innovations in estimating the prediction of complex systems~\citep{brunton2019data}. With a deeper understanding of neural networks, it may be possible to generalize the non-intrusive approach to another level of accuracy. Furthermore, physics-informed models~\citep{raissi2019physics} may accelerate learning from data and add interpretability to existing approaches.  
This framework can also be combined with distributed systems to enable parallelism for extreme-scale high performance computing.

\section*{Acknowledgments}
We would like to thank Kevin Carlberg, Steve Brunton and Youngsoo Choi for valuable discussions. ZB acknowledges support by the U.S. Department of Energy, Office of Science, under Award Number DE-AC02-05CH11231. ZB and LP acknowledge support for preliminary work from Sandia National Laboratories. Any subjective views or opinions that might be expressed in the paper do not necessarily represent the views of the U.S. Department of Energy or the United States Government. Sandia National Laboratories is a multimission laboratory managed and operated by National Technology \& Engineering Solutions of Sandia, LLC, a wholly owned subsidiary of Honeywell International Inc., for the U.S. Department of Energy’s National Nuclear Security Administration under contract DE-NA0003525.

\bibliography{references} 
\bibliographystyle{unsrtnat}

\clearpage 
\appendix
\renewcommand\thefigure{\thesection.\arabic{figure}}  
\renewcommand\theequation{\thesection.\arabic{equation}}  
\renewcommand\thetable{\thesection.\arabic{table}} 
\section{Regression models and tuning} \label{app}
\setcounter{figure}{0} 
\setcounter{equation}{0} 
\setcounter{table}{0} 
\subsection{Special regression models}\label{app1}
In this section, we detail two non-transitional machine learning models, VKOGA and SINDy, in addition to the SVR, kNN, random forest and boosting.
\subsubsection{Vectorial kernel orthogonal greedy algorithm (VKOGA)} 
Developed from kernel-based methods, VKOGA makes the assumption that a common subspace can be used for every component of vector-valued response. With such, it significantly reduces the training and evaluation time when the number of kernels in the resulting regression model is large. VKOGA constructs the regression model
\begin{equation}
\func(\mystate) = \sum_{i=1}^{n_{\text{VKOGA}}}{\boldsymbol \alpha_i} K(\mystate_i,\mystate),
\end{equation}
where $K:\RR{\nstate_x}\times\RR{\nstate_x} \rightarrow\RR{}$ denotes a kernel function, for each individual regression model $h_i$, $i=1,\cdots,\nstate_x$, and $\alpha_i \in \RR{\mystate_x}, i=1,\cdots,\nstate_x$ are vector-valued basis functions.
VKOGA uses a greedy algorithm to compute kernel functions $K(\bar{\mystate}, \cdot)$. The greedy algorithm determines kernel centers from $\Omega = {\bar{\mystate}^1, \cdots, \bar{\mystate}^{n_{train}}}$. Let the initial state $\Omega_0=\emptyset$, and then at stage $m$, choose
\begin{equation}
    \mystate_m = \argmax_{\mystate \in \Omega \backslash \Omega_{m-1}} |\langle \func, \phi_x^{m-1}\rangle|, 
\end{equation}
where $\phi_x^{m-1}$ denotes the orthogonal remainder of $K(\mystate, \cdot)$ with respect to the reproducing kernel Hilbert space spanned by ${K(\mystate_1, \cdot)}, \cdots, K(\mystate_{m-1}, \cdot)$. Then the kernel centers are updated $\Omega_m = \Omega_{m-1} \cup {\mystate_m}$. At the next step, the basis function $\boldsymbol\alpha_i$, $i=1, \cdots, n_{\text{VKOGA}}$ are determined by a least-squares approximation to fit the training data.  
In the numerical experiments, we apply the Gaussian RBF kernel $K(\bar{\mystate}^i, \bar{\mystate}^j) = exp(-\gamma \|\bar{\mystate}^i, \bar{\mystate}^j\|)$, where the hyperparameter is the number of kernel functions used. 

 \subsubsection{Sparse identification of nonlinear dynamics (SINDy)} 
 The sparse identification of nonlinear dynamics (SINDy) method~\citep{Brunton2016pnas} identifies nonlinear dynamical system from measurement data, seeking to approximate the vector field $\velocity$ by a generalized linear model. This architecture avoids overfitting by identifying a parsimonious model, which is more explainable or predictable than black-box machine learning. Although SINDy was originally devised for continuous-time dynamics, it can be extended to discrete-time systems. In particular, SINDy yields a model
 \begin{equation}
 \func(\mystate, \theta) = \sum_{i=1}^{n_\text{{SINDy}}} p_i(\mystate)\theta_i ,
 \end{equation}
where $p_i:\RR{N_x}\rightarrow\RR{}$, $i = 1, \cdots, n_{\text{SINDy}}$ denotes a "library" of candidate basis functions. 
The relevant terms that are active in the dynamics are solved using sparse regression that penalizes the number of functions in the library $\boldsymbol P(\mystate)$:
\begin{equation}
    \theta_i = \argmin_{\theta'_i} \|\func-\boldsymbol P(\mystate)\theta'_i \|_2 + \alpha\|\theta'_i\|_1,
\end{equation}
where $\|\cdot\|_1$ promotes sparsity in the coefficient vector $\theta_i$ and the parameter $\alpha$ balances low model complexity with accuracy.
The least absolute shrinkage and selection operator (LASSO)~\citep{tibshirani1996lasso} or the sequential threshold least-squares (STLS)~\citep{Brunton2016pnas} can be used to determine a sparse set of coefficients ($\theta_1, \cdots, \theta_{n_{\text{SINDy}}}$).
In the numerical experiments, we consider linear and quadratic functions in the library, i.e., $p_i \in \{x_1, \cdots, x_{N_x}, x_1x_1, x_1x_2,\cdots, x_{N_x}x_{N_x}\}$. Hyperparameters in this approach consist of the prescribed basis functions. 

\ref{tab:compMLmethods} summarizes the computational complexities of all the considered regression models. Note that the hyperparameter varies in different models, i.e. $N_\text{trees}$ denotes the number of decision tress in random forest, $N_\text{learners}$ denotes the number of weak learners in the boosting method, $N_\text{functions}$ is the numner of kernel functions in VKOGA, and $N_\text{bases}$ represents the number of bases in SINDy with 2$^{nd}$ order polynomials.   

\begin{table}[ht]
\begin{center}
\caption{Theoretical complexity per time step, where $\nstateRed$ represents the subspace dimension, $\nparams$ represents the parameter dimension, and $\ntrain$ represents the number of training points.} 
\label{tab:compMLmethods}
 {%
\begin{tabular}{cccc}
\toprule
\textbf{Method}  &   \textbf{Complexity}  & \textbf{Comments} \\
\midrule
 SVR 2  &  O($(\nstateRed+ \nparams) \nstateRed \ntrain$) &   \\

SVR 3 & O($(\nstateRed+ \nparams) \nstateRed \ntrain$) &  \\

SVR rbf  & O($(\nstateRed+ \nparams) \nstateRed \ntrain$) &  \\

Random Forest  & O($\nstateRed  N_\text{trees} \ntrain \log(\ntrain)$) & $N_\text{trees}$: number of decision trees  \\ 

Boosting  & O($\nstateRed N_\text{learners}$) &  $N_\text{learners}$: number of weak learners \\

kNN   &  O($(\nstateRed+ \nparams)  \nstateRed \ntrain +  K \nstateRed \ntrain$) &  $K$: number of nearest neighbors\\

VKOGA  & O($(\nstateRed+ \nparams) N_\text{functions}$) & $N_\text{functions}$:  number of  kernel functions \\

SINDy2  & O($\nstateRed N_\text{bases}$) & $N_\text{bases}$: number of  bases, $N_\text{bases} < \nstateRed ^2$\\
\bottomrule
\end{tabular}}
\end{center}
\end{table}

\subsection{Cross-validation and hyperparameter tuning} \label{app2}
For each regression model proposed in Section~\ref{sec3.2}, we use cross-validation to prevent overfitting and select hyperparameters.

\begin{figure}\label{AppFig3}
	\centering
	\begin{overpic}[width = 0.99\textwidth]{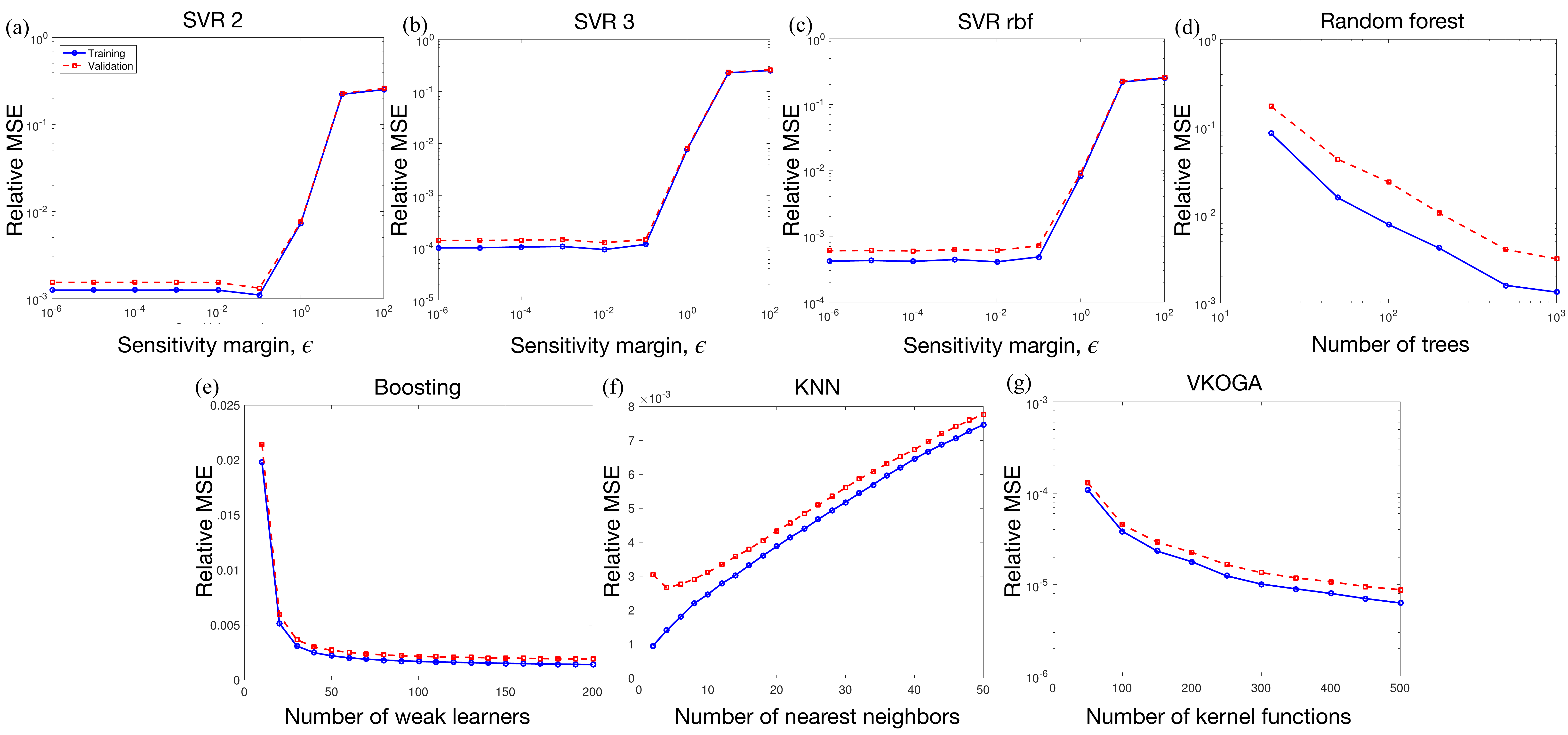}
	\end{overpic}
	\caption{Hyperparameter selection for all considered regression models. Relative mean squared errors are reported: (a) SVR2, (b) SVR3, (c) SVR rbf, (d) random forest, (e) boosting, (f) kNN, (g) VKOGA. The training and validation set correspond to the default size: $\ntrain = 1000$ and $\nvalidation = 500$. Blue curves represent training errors and red curves represent validation errors.}   
\end{figure}

\begin{figure}[t]\label{AppFig4}
	\centering
	\begin{overpic}[width = 0.99\textwidth]{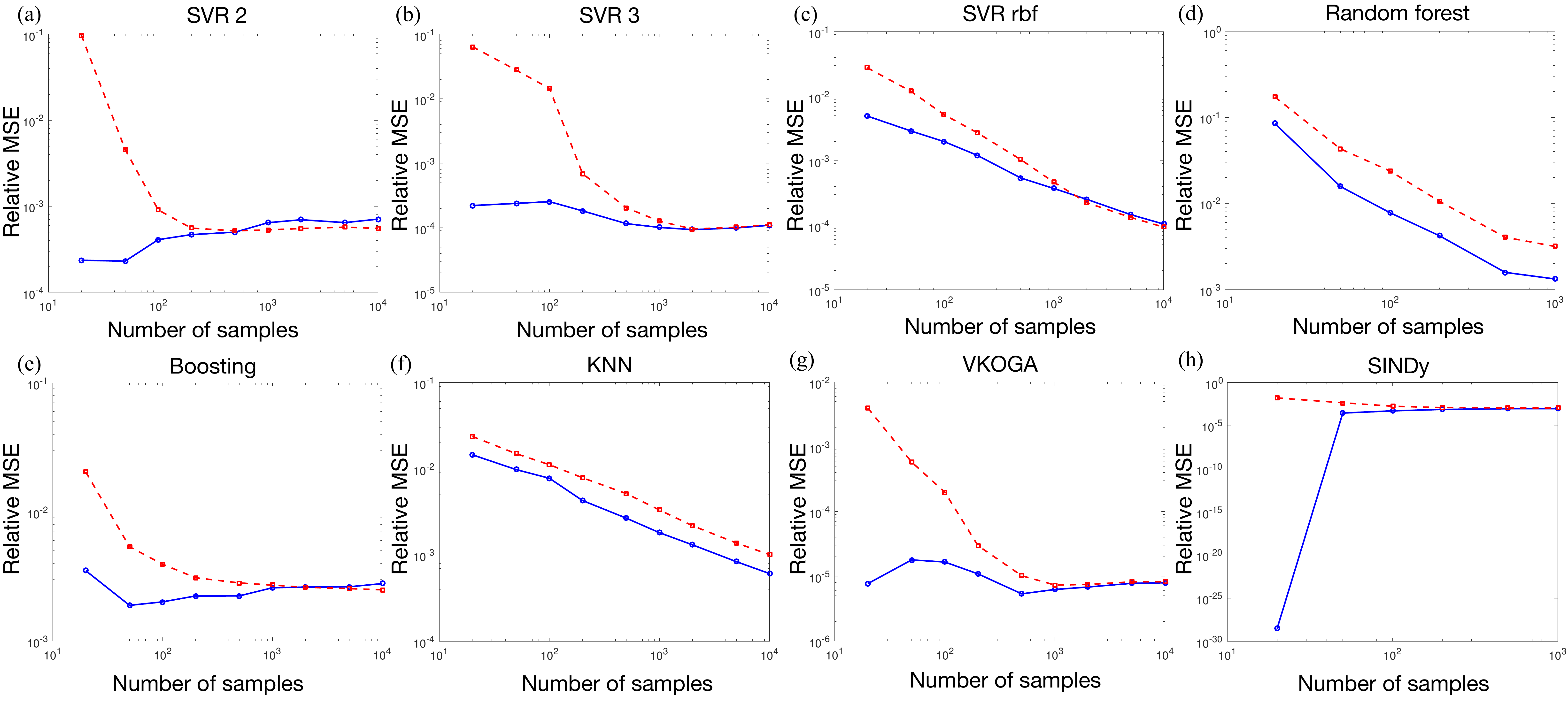}
	\end{overpic}
	\caption{Relative mean squared error vs. training size: (a) SVR2, $\epsilon=10^{-4}$, (b) SVR3, $\epsilon=10^{-4}$, (c) SVR rbf, $\epsilon=10^{-4}$, (d) random forest, (e) boosting, (f) kNN, (g) VKOGA. Note that the validation set is fixed, $\nvalidation = 500$. Blue curves represent training errors and red curves represent validation errors.}
\end{figure}
We report the hyperparameter selection of each model for the 2D convection-diffusion equation in Figure~\ref{AppFig3}.
Figure~\ref{AppFig3}(a) shows that with the sensitivity margin $\epsilon \le 0.1$, both the training error and validation errors are very small, and thus we select the sensitivity margin $\epsilon=0.1$ for SVR2. Similarly, $\epsilon=10^{-3}$ is selected for SVRrbf. 
Figure~\ref{AppFig3}(d) shows that within 40 trees, the validation error of the random forest model is still relatively large.  Moreover, there is a big gap between the training error and test error that implies the random forest model may suffer from overfitting for the given data. We select $N_\text{trees}=15$ in this study.
As shown in Figure~\ref{AppFig3}(e), the validation error in Boosting approaches the minimum at $N_\text{learners} = 40$, and thus we select $40$ weak learners.
Figure~\ref{AppFig3}(f) shows that by choosing the number of neighbors $K \approx 4$, the validation error is minimized.  Thus, $K = 4$ is a good hyperparameter to balance bias and variance. 
In Figure~\ref{AppFig3}(g) we observe that the more kernel functions we use, the smaller value of training and validation error we obtain. In this study, we select 500 kernel functions in VKOGA.

We examine the performance of the models as an increasing size of the training sample in Figure~\ref{AppFig4}.
Figure~\ref{AppFig4}(a) shows that with $200$ training instances, the SVR2 model has very small validation error, while SVR3 model requires a bigger training set. For SVRrbf, the more training instances we have, the smaller validation error we obtain. This is because the resolution of the radial basis function is determined by the density of input parameter space.
Figure~\ref{AppFig4}(d) shows that the more data are input, the less the relative error. Intuitively, more data give higher density in the input parameters space; and thus each leaf of a decision tree provides finer coverage for test data. Similarly, for the kNN model as in~Figure~\ref{AppFig4}(f), close neighbors give better representation of test data. As a result, more data can effectively reduce the variance and solve the overfitting issue. 
Figure~\ref{AppFig4}(e) shows that with approximately $1000$ training samples, the training error and validation error begin to approach $2e\text{-}3$. This implies that a relative large number of data are required in order to constrain the variance of the boosting method.
Figure~\ref{AppFig4}(g) shows that with $1000$ training instances, the training and validation errors begin to converge for VKOGA, while SINDy (2$^{nd}$ order polynomials) only requires a small number of training instances to get a very accurate prediction, as seen in~\ref{AppFig4}(h). Similarly, we select the following hyperparameters: $\epsilon=10^{-4}$ for SVR2, $\epsilon=10^{-5}$ for SVR3, $\epsilon=10^{-3}$ for SVRrbf, $N_\text{trees} = 15$ for random forest, $N_\text{learners} = 40$ for boosting, $K=6$ for $k$-nearest neighbours, and $500$ kernel functions for VKOGA in the 1D Burgers' implementation.

\begin{figure}[t]
	\centering
	\vspace{0.4in}
	\begin{overpic}[width = 0.455\textwidth]{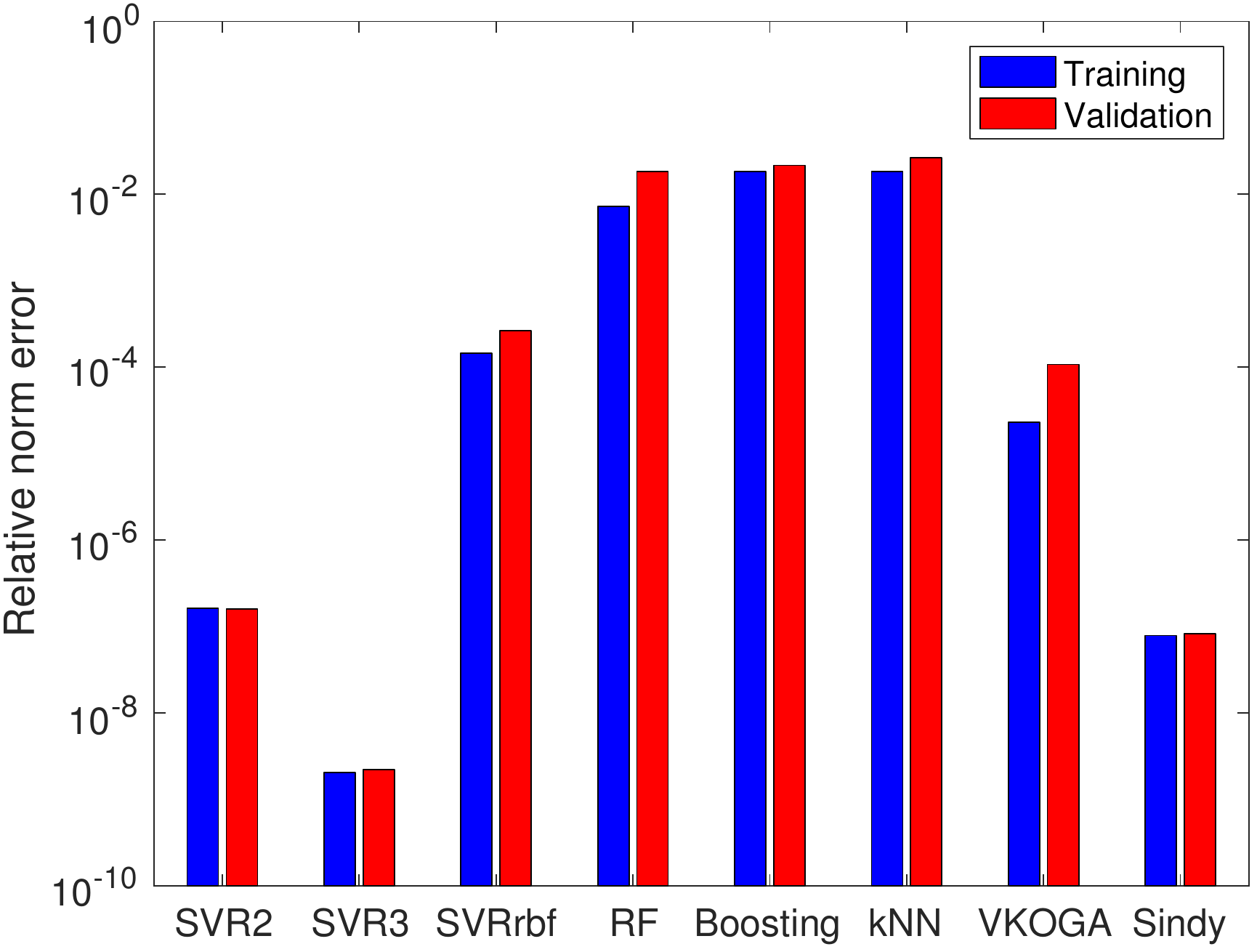}
	\put(-0.3,77){(a)}
	\end{overpic}
	\begin{overpic}[width = 0.45\textwidth]{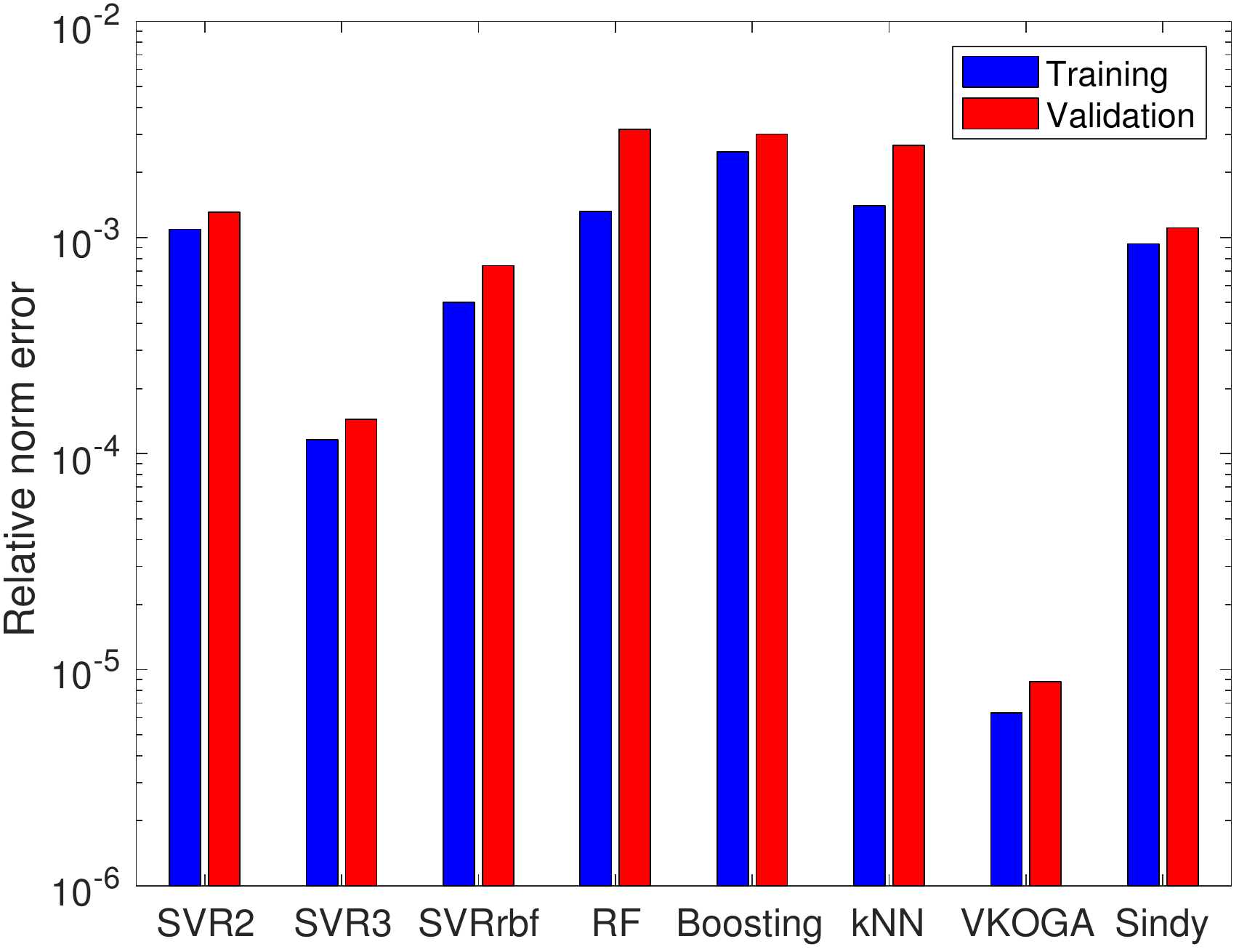}
	\put(-0.4,78){(b)}
	\end{overpic}
	\caption{Relative errors with $\ntrain=1000$ and  $\nvalidation = 500$: (a) 1D Burgers' equation; and (b) 2D convection-diffusion equation. Blue bars represent training errors and red bars represent validation errors.}
	\label{FigOverall}
\end{figure}
Figure~\ref{FigOverall} summarizes the overall performance of the above models for the 1D Burgers' and 2D convection-diffusion equation.
All the proposed models are validated using the selected hyperparameters. We notice that SVR with the kernel of 3$^{nd}$ order polynomials outperforms the other models for the 1D Burgers' equation. This can be related to the discretization structure of quadratic nonlinearties in the PDE solver. VKOGA has the a better reperformance than RF, Boosting, kNN, with a relative error less than $1e{\text -}4$. SINDy reaches a relative error of $3e{\text -}8$. For the 2D convection-diffusion problem, SVR models again perform better than RF, Boosting and kNN, and SVR3 outperforms the other two kernel functions. VKOGA has the best accuracy with relative error $\approx 1e{\text -}5$, while SINDy reaches a relative error of $1e{\text -}3$. Note that to be consistent, we select fixed training and validation sample sizes for all the considered regression models. However, as illustrated in Figure~\ref{AppFig4}(h), SINDy can approach the same level of accuracy with a small number of training and validation samples.

\end{document}